\documentclass{article}

\usepackage[final,main]{neurips_2026}

\usepackage[utf8]{inputenc} 
\usepackage[T1]{fontenc}   
\usepackage{hyperref}       
\usepackage{url}            
\usepackage{booktabs}       
\usepackage{amsfonts}       
\usepackage{nicefrac}      
\usepackage{microtype}     
\usepackage{xcolor}        
\usepackage{multirow}
\usepackage{pifont}
\usepackage{algorithm}
\usepackage{algpseudocode}
\usepackage{graphicx}
\usepackage{amsmath}
\usepackage{comment}

\usepackage[nameinlink,capitalize]{cleveref} 
\usepackage{tikz}
\usetikzlibrary{arrows.meta,positioning,shapes.geometric,fit,backgrounds,calc}
\usepackage{float}
\usepackage{placeins}
\usepackage{amsthm}
\theoremstyle{plain}
\newtheorem{theorem}{Theorem}
\newtheorem{proposition}{Proposition}

\newtheorem{definition}{Definition}
\newtheorem{remark}{Remark}

\newcommand\Rset{{\mathbb{R}}}
\newcommand\Nset{{\mathbb{N}}}

\title{Ensuring Deployment-Time Safety of Neural Network Controlled Systems via Localized Certificate Repair}

\author{%
  \normalfont
  \begin{tabular}{@{}c@{}}
    \textbf{Xiaoyang Lv}\textsuperscript{1},
    \textbf{Peixin Wang}\textsuperscript{1}\thanks{Corresponding author. Email address: pxwang@sei.ecnu.edu.cn},
    \textbf{Jianhao Bai}\textsuperscript{1},
    \textbf{Bai Xue}\textsuperscript{2},
    \textbf{Min Zhang}\textsuperscript{1}\footnotemark[1],
    \textbf{Luke Ong}\textsuperscript{3}
    \\[0.4em]
    \textsuperscript{1}\,
    Shanghai Key Laboratory of Trustworthy Computing, East China Normal University
    \\
    \textsuperscript{2}\,
    KLSS, Institute of Software, Chinese Academy of Sciences
    \\
    \textsuperscript{3}\,
    College of Computing and Data Science, Nanyang Technological University
 \\
  \end{tabular}%
}

\begin{document}

\maketitle

\begin{abstract}
Barrier certificates provide formal safety guarantees for neural network controlled systems by ensuring that system trajectories avoid unsafe regions over time. However, after deployment, previously unseen obstacles or constraint changes may introduce new unsafe regions, invalidating the original certificate and potentially leading to unsafe behavior.
We address this setting with \emph{adaptive piecewise barrier certificates}, which preserve the original certificate and policy in unchanged regions while introducing localized updates around newly emerged unsafe regions. The central question is how to perform such updates efficiently during system execution without retraining from scratch.
Our key idea is to cast deployment-time adaptation as a localized certificate repair problem and reduce it to an optimization task that leverages the learned structure of previous barrier certificates, enabling fast updates without global retraining. We further design a staged repair framework that progressively transitions from localized repair to joint policy–certificate updates and, when necessary, localized retraining. 
Empirical results on four benchmarks with five new obstacles show that our method succeeds in all cases and completes within tens of seconds on average, whereas retraining from scratch fails in most cases. On instances where retraining succeeds, our method is \textbf{17.3$\times$} faster on average. It incurs only \textbf{16.2\%} performance degradation relative to the original policy.

\end{abstract}

\section{Introduction}\label{sec:intro}
Neural network controlled systems (NNCSs), particularly those based on deep reinforcement learning (DRL), are increasingly deployed in real-world applications such as robotic manipulation~\cite{kalashnikov2018scalable}, locomotion~\cite{hwangbo2019learning}, and autonomous driving~\cite{kiran2021deep}. In these safety-critical settings, ensuring reliable behavior beyond training phases is essential, as violating state or input constraints can lead to severe consequences~\cite{dulac2019challenges}. Barrier certificates (BCs) provide a principled approach for formal safety guarantees by ensuring that system trajectories avoid unsafe regions over time~\cite{prajna2004safety,ames2019control}. However, such guarantees are typically established for a fixed set of obstacles and constraints. After deployment, newly emerged obstacles may introduce unsafe regions unseen during training, invalidating the original certificate even for a previously certified system. This raises a fundamental challenge: how to maintain formal safety guarantees in the presence of newly introduced unsafe regions at deployment time.

A natural solution is to retrain the controller and certificate in the updated environment as commonly done in safe reinforcement learning and control synthesis~\cite{garcia2015comprehensive,cheng2019end}. However, global retraining is often too costly for deployment-time budgets ~\cite{dulac2019challenges}. Alternatively, runtime verification and shielding enforce safety through online monitoring and intervention ~\cite{alshiekh2018safe,jansen2020safe,banerjee2024dynamic}. 
While shielding approaches provide strong safety guarantees, they are particularly suited to transient environmental changes where obstacles may appear and soon disappear.
For persistent localized changes, repeated interventions around the same obstacle can incur substantial cumulative overhead.
A representative setting arises in autonomous mining and inspection,
where systems repeatedly operate in largely stable environments
with occasional but persistent localized changes~\cite{jacobson2021localizes,
mansouri2020deploying}.

To address these limitations, we seek an approach that restores safety efficiently at deployment time while avoiding both the cost of global retraining and the overhead of continuous runtime monitoring and intervention. Our approach is based on the observation that environment changes typically affect only a localized subset of the state space, so that safety can be recovered by modifying the barrier certificate only where necessary. This motivates a one-time certified repair, allowing the localized policy and certificate to be reused upon subsequent encounters without further repair.
Building on the observation, we consider neural barrier certificates~\cite{abate2021fossil}, where the certificate is represented by a neural network, and propose a staged adaptation strategy centered on localized certificate repair. 
Inspired by neural network repair ~\cite{sotoudeh2021provable,refaeli2022minimal,majd2021local,dong2021towards}, the method first repairs the original certificate in affected regions by reusing learned certificate structure, which can be formulated as a tractable linear program. When such repair is insufficient, it incrementally increases flexibility by jointly updating the policy and the certificate through a localized bilevel optimization. As a final fallback, it performs retraining restricted to the affected regions only, avoiding unnecessary recomputation elsewhere.
This design provides a principled middle ground between retraining and runtime monitoring: it restores persistent safety guarantees through targeted, one-time updates, while maintaining efficiency by reusing the structure of the original neural certificate and policy.

Our main contributions are as follows:
\begin{itemize}
    \item We introduce \textbf{adaptive piecewise barrier certificates}
    (\cref{df:apwbc}) for deployment-time safety repair, preserving the original policy and certificate in unaffected regions while composing them with locally repaired counterparts.
    We establish interface compatibility conditions that guarantee global safety across region transitions (\cref{theorem:apwbc}).

    \item We develop a localized repair pipeline that reuses the structure of pre-trained certificates.
    We reduce continuous BC constraints to finite-grid constraints (\cref{thm:LipBC}).
    The pipeline prioritizes certificate repair under a fixed policy, followed by local policy adaptation when necessary, with localized retraining as a fallback
    (\cref{thm:mainstage1_success,thm:mainstage2_success}).

    \item We implement and evaluate our approach on four benchmarks with five newly introduced obstacles.
    Our method restores global safety within seconds, achieving an average \textbf{17.3$\times$} speedup over successful global retraining runs, while preserving nominal performance with an average reward degradation of $16.2\%$.
\end{itemize}

\section{Related Work}\label{sec:related}
\noindent\textbf{Barrier Certificates for Safety Verification.}
Barrier certificates verify safety of dynamical systems by establishing forward invariance of safe sets without explicit reachability analysis~\cite{prajna2004safety}. They have been extensively studied in control theory, including the development of control barrier functions for enforcing safety in nonlinear systems~\cite{ames2019control}. 
A recent work use barrier certificates to provide safety guarantees in prescribed dynamic obstacles~\cite{mazouz2026stochastic}.
More recently, barrier certificates have been combined with learning-based methods, where neural networks are used to represent certificates or controllers, enabling scalable safety verification for high-dimensional systems~\cite{cheng2019end,dawson2023safe}. 
However, existing barrier certificate methods typically assume a fixed environment and focus on offline synthesis and verification. They do not address the problem of maintaining safety when new obstacles or constraints are introduced after deployment, which may invalidate previously certified guarantees.

\noindent\textbf{Safe Reinforcement Learning.}
Safe reinforcement learning aims to learn policies that satisfy safety constraints during training or execution~\cite{garcia2015comprehensive}. A common approach is to incorporate constraints into the learning objective or to synthesize safe controllers jointly with certificates, including methods based on barrier functions~\cite{cheng2019end}. While effective in principle, these approaches typically rely on retraining when the environment or constraints change, which can be computationally expensive and impractical under deployment-time constraints~\cite{dulac2019challenges}. In contrast, our work focuses on efficient safety restoration without global retraining.

\noindent\textbf{Runtime Verification and Shielding.}
Runtime verification and shielding ensure safety through online monitoring and intervention~\cite{alshiekh2018safe,jansen2020safe,banerjee2024dynamic}.
These methods are particularly suited to transient environmental changes, where obstacles may appear and disappear.
For persistent localized changes, however, repeated interventions around the same obstacles can accumulate substantial overhead.
In contrast, our approach performs one-time localized repair and verification, allowing the certified policy and certificate to be reused across subsequent executions.

\noindent\textbf{Neural Network Repair.}
Neural network repair techniques aim to correct undesirable behaviors of trained models through minimal modifications, without retraining from scratch~\cite{sotoudeh2021provable,refaeli2022minimal,majd2021local,dong2021towards}. These methods have been studied in the context of formal verification and robustness, focusing on enforcing local correctness properties while preserving overall model behavior. Our work is inspired by this line of research, but differs in that we develop repair techniques for safety certificates in dynamical systems and integrate them into a deployment-time safety framework.

\section{Preliminaries}\label{sec:pre}
\subsection{Neural Network Controlled Systems}
An NNCS can be modeled as a tuple
$M=(S,S_0,A,f,\pi,R,S_u)$, where $S \subseteq \mathbb{R}^n$ is the (continuous) state space and $S_0 \subseteq S$ is a set of admissible initial states.
At each discrete time step $t \in \mathbb{N}_0$, the controller chooses an action $a_t \in A$ from an action space $A \subseteq \mathbb{R}^m$ according to a control policy $\pi : S \to A$, implemented by a deep neural network (DNN) via deep reinforcement learning.
Concretely, the system evolves according to the following dynamics
\begin{equation}
    x_{t+1} = f(x_t,a_t), \qquad f : S \times A \to S,
\end{equation}
where $x_t \in S$ denotes the state at time $t$. Consequently, a reward $r_t(x_t,a_t,x_{t+1})$ is obtained by the reward function $R:S\times A\times S\rightarrow \Rset$.
Safety requirements are specified by an \emph{unsafe set} $S_u \subseteq S$ such that $S_0 \cap S_u = \emptyset$ (equivalently, a safe set is $S \setminus S_u$).

\noindent \textbf{Trajectories.} 
Given an NNCS $M$ and an initial state $s_0 \in S_0$, a sequence of states generated during interaction with the environment is called a \textit{trajectory}, denoted as $\omega = \{x_t\}_{t \in \mathbb{N}_0}$. 
To simplify notations, we denote by $\omega_t$ the \textit{t}-th element of $\omega$, i.e., $\omega_t=x_t$, and by $\Omega_{M} (S_0)$ the set of all trajectories from $S_0$ in the NNCS $M$. Moreover, we define 
$\mathcal H(\Omega_M(X))
:=
\bigcup_{\omega\in\Omega_M(X)}
\{\omega_t\in S \mid t\in\mathbb N_0\}$, which collects all states appearing along trajectories in \(\Omega_{M} (X)\).

\subsection{Barrier Certificates and Safety Guarantees}\label{sec:bc-intro}
We briefly review barrier certificates for discrete-time systems and the safety guarantee~\cite{prajna2004safety,ames2019control,xue2024sufficient}. 

\begin{definition}[Barrier certificates]\label{df:bc}
A barrier certificate for an NNCS $M$ is a function $B : S \to \mathbb{R}$ if it satisfies that for some constant $\lambda\in [0,1]$:
\begin{align}
    &\text{(Initial condition)} && B(x) \le 0, && \forall x \in S_0, \label{eq:bc-init}\\
    &\text{(Separation)} && B(x) > 0, && \forall x \in S_u, \label{eq:bc-sep}\\
    &\text{(Non Positivity)} && B\bigl(f(x,\pi(x))\bigr)-\lambda\cdot B(x) \le 0, && \forall x \in S\setminus S_u. \label{eq:bc-inv}
\end{align}
\end{definition}
Condition~\eqref{eq:bc-init} ensures that all admissible initial states satisfy the safety condition.
Condition~\eqref{eq:bc-sep} enforces a strict separation between safe and unsafe states.
Condition~\eqref{eq:bc-inv} guarantees that the barrier value does not increase along system trajectories, preventing transitions from safe to unsafe regions.

\begin{theorem}[Safety Guarantees by Barrier Certificates]\label{thm:safety-BCs}
If there exists a barrier certificate $B$ for an NNCS $M$ satisfying~\eqref{eq:bc-init}--\eqref{eq:bc-inv}, then any trajectory $\omega=\{x_t\}_{t\in\mathbb{N}_0}\in\Omega_{M}(S_0)$ with $x_0\in S_0$ satisfies $x_t \notin S_u$ for all $t\in\mathbb{N}_0$. Consequently, $M$ is safe w.r.t. $S_u$.
\end{theorem}

\noindent \textbf{Neural barrier certificates (NBC).}
Barrier certificates can be parameterized by neural networks~\cite{abate2021fossil,peruffo2021automated}, providing strong expressive power for high-dimensional systems. Under the CEGIS paradigm~\cite{abate2018counterexample}, NBC synthesis alternates between learning a candidate certificate and validating it against the barrier conditions in~\cref{eq:bc-init}--\cref{eq:bc-inv}, with counterexamples used for refinement~\cite{vzikelic2023learning,zhi2024robustness}. 
In this work, we assume a nominal policy $\pi$ is given, along with an NBC $B$ that has been synthesized and validated in a nominal environment, and focus on repairing $B$ under deployment-time changes.

\noindent\textbf{Assumptions.} We assume the system dynamics, policy, and neural barrier certificates are all Lipschitz continuous. Furthermore, we suppose the state space $S$ is compact.
 All the assumptions are commonly used in the literature~\cite{vzikelic2023learning,zhi2024robustness,zhi2024unifying}.

\section{Deployment-time Safety Guarantee}\label{sec:framework}

\subsection{Motivation and Problem Statement}\label{sec:problem}

\begin{figure}[!htbp]
    \centering
    \begin{minipage}[t]{0.53\linewidth}
        \centering
        \includegraphics[width=\columnwidth,trim={20pt 16pt 8pt 25pt},
  clip]{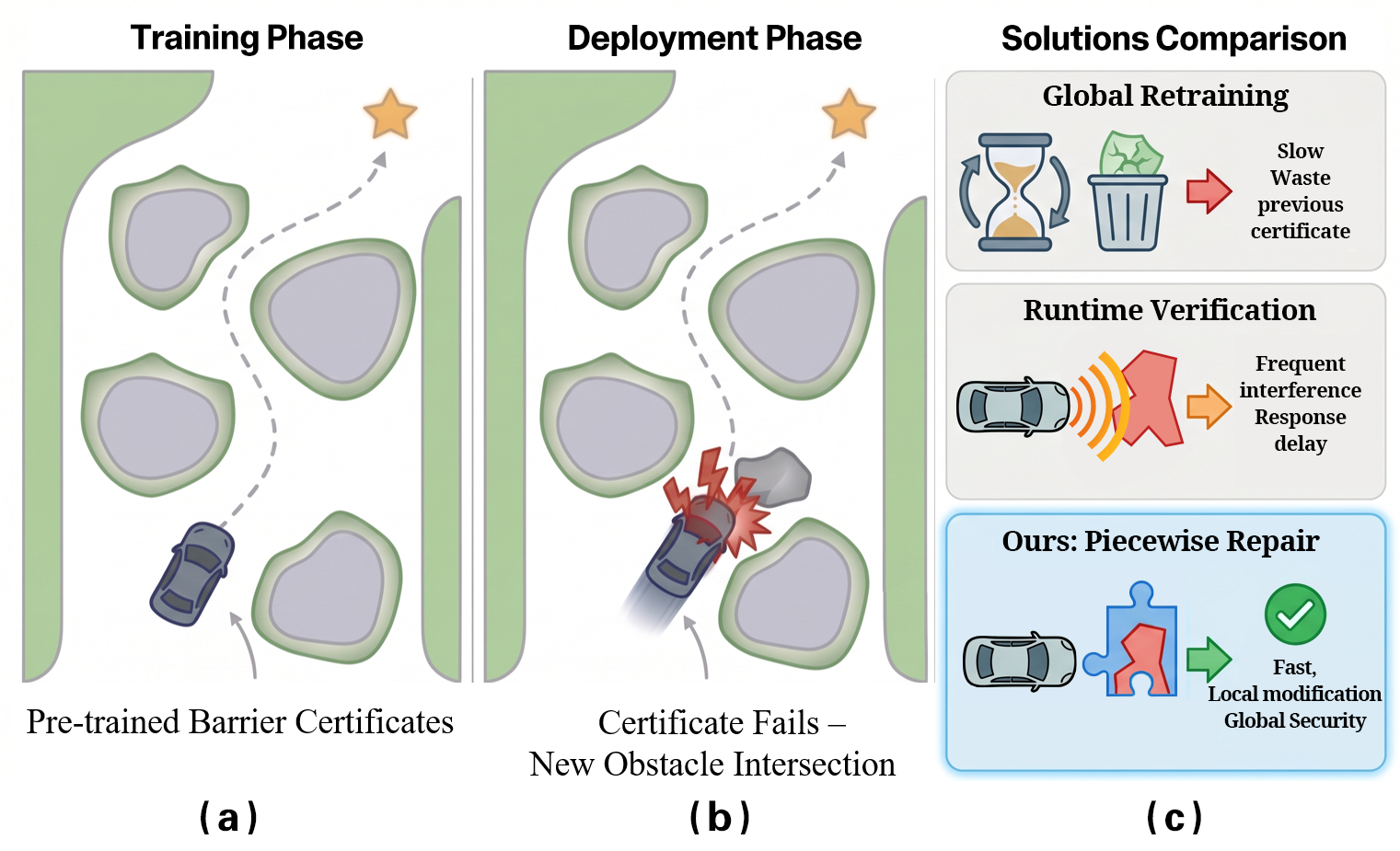}
        
        \caption{
            Problem Statement and Methods Comparison
        }
        \label{fig:problem_statement}
    \end{minipage}
    \hfill
    \begin{minipage}[t]{0.43\linewidth}
        \centering
        \includegraphics[width=\columnwidth]{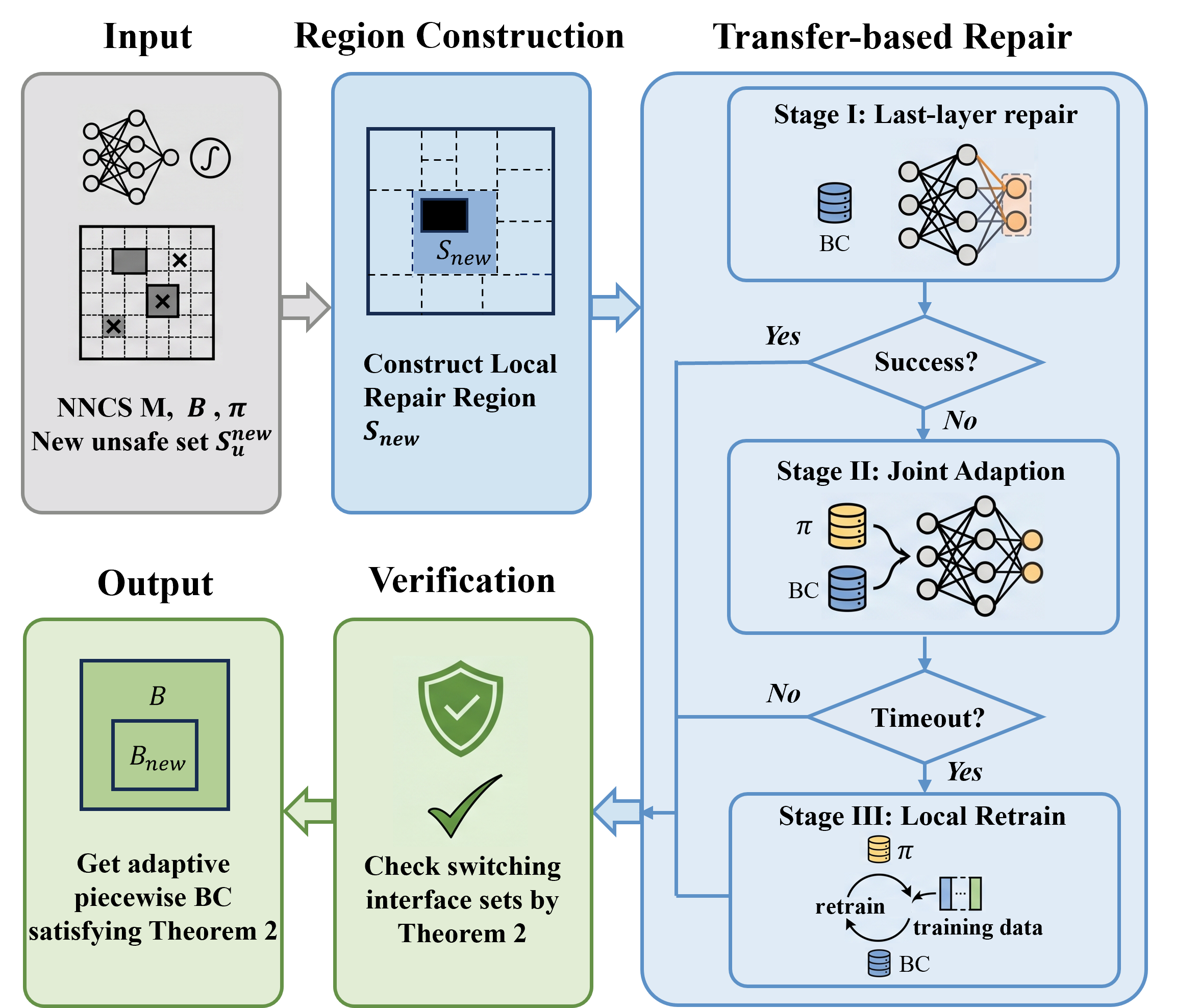}
       
        \caption{
            Overview of the proposed local repair framework.
        }
        \label{fig:repair_framework}
    \end{minipage}
    
\end{figure}

Consider an NNCS $M=(S,S_0,A,f,\pi,R,S_u)$ with a control policy $\pi$ and a barrier certificate $B$ that guarantees safety with respect to the unsafe set $S_u$, as shown in~\cref{fig:problem_statement}(a). After deployment, the environment introduces a new unsafe set $S_u^{\mathrm{new}}\subseteq S$ with $S_u\cap S_u^{\mathrm{new}}=\emptyset$, leading to an updated unsafe set $\hat{S}_u := S_u \cup S_u^{\mathrm{new}}$. Consequently, $M$ may enters the new unsafe set $S_u^{\mathrm{new}}$ (see \cref{fig:problem_statement}(b)). This loss of safety is caused by two reasons. First, the policy $\pi$ is not trained against $S_u^{\mathrm{new}}$ and can drive the system into this region. Second, the barrier certificate $B$, synthesized for $S_u$, may fail to certify safety with respect to $\hat{S}_u$, e.g., the separation condition can be violated on $S_u^{\mathrm{new}}$. 

\noindent\textbf{Existing approaches and limitations.}
\emph{Global retraining} updates both the policy and barrier certificate, but incurs high computational costs, and must be repeated for new unsafe regions.
\emph{Runtime verification and shielding} handle transient environmental changes through online monitoring and intervention, but repeated interventions around persistent obstacles can accumulate substantial overhead.
For persistent localized changes, we instead seek one-time deployment-time repair to restore safety and reuse the certified system across subsequent executions
(\cref{fig:problem_statement}(c)).

\noindent \textbf{Problem Statement.}
Given a newly introduced unsafe set $S_u^{\mathrm{new}}$ after deployment, leading to the updated unsafe set $\hat{S}_u := S_u \cup S_u^{\mathrm{new}}$, the goal is to restore safety with respect to $\hat{S}_u$, i.e., $x_t\notin \hat{S}_u$ for all time $t\in\Nset_0$. This should be achieved efficiently by preserving the original policy and certificate whenever possible.

\subsection{Safety by Adaptive Piecewise Barrier Certificates}\label{sec:Theoretical}

To restore safety with respect to the updated unsafe set $\hat{S}_u = S_u \cup S_u^{\mathrm{new}}$ after deployment-time changes, we avoid global re-certification and instead perform localized updates. Specifically, we partition the state space $S$ into an unaffected region $S_{\mathrm{origin}}$ and a localized repair region $S_{\mathrm{new}}$ (see~\cref{sec:BC transfer}) such that $S = S_{\mathrm{origin}}\cup S_{\mathrm{new}}$ and $S_{\mathrm{origin}}\cap S_{\mathrm{new}}=\emptyset$. Moreover, $S_u\subseteq S_{\mathrm{origin}}$ and $S_u^{\mathrm{new}}\subseteq S_{\mathrm{new}}$ and $S_0 \subseteq S_{\mathrm{origin}}$. On $S_{\mathrm{origin}}$, the original policy $\pi$ and barrier certificate $B$ are retained, while on $S_{\mathrm{new}}$, a localized policy $\pi_{\mathrm{new}}$ is used.\footnote{$\pi_{\mathrm{new}}$ is set to be $\pi$ and then updated if necessary to restore system safety, see \cref{sec:solutions}. } The resulting system is described by $\hat{M}=(S,S_0,A,f,\hat{\pi},R,\hat{S}_u)$, where the closed-loop policy $\hat{\pi}$ is defined piecewise over the two regions, i.e., $\hat{\pi}=\pi(x)$ when $x\in S_{\mathrm{origin}}$ and $\hat{\pi}=\pi_{\mathrm{new}}(x)$ when $x\in S_{\mathrm{new}}$.

To characterize the cross-region transitions, we introduce the following switching interface sets:
\begin{align}
\Gamma_{\mathrm{in}}^{-}
&:= \{x \mid x\in \mathcal H (\Omega_{\hat{M}}(S_0))\cap S_{\mathrm{origin}},\ f(x,\pi(x))\in S_{\mathrm{new}}\}, \label{set1} \\
\Gamma_{\mathrm{in}}^{+}
&:= \{f(x,\pi(x)) \mid x\in \Gamma_{\mathrm{in}}^{-}\}, \label{set2} \\
\Gamma_{\mathrm{out}}^{-}
&:= \{x \mid x\in \mathcal H (\Omega_{\hat{M}}(\Gamma_{\mathrm{in}}^{+}))\cap S_{\mathrm{new}},\ f(x,\pi_{\mathrm{new}}(x))\in S_{\mathrm{origin}}\}, \label{set3} \\
\Gamma_{\mathrm{out}}^{+}
&:= \{f(x,\pi_{\mathrm{new}}(x)) \mid x\in \Gamma_{\mathrm{out}}^{-}\}. \label{set4}
\end{align}
Here $\Gamma_{\mathrm{in}}^{-}$ and $\Gamma_{\mathrm{in}}^{+}$ describe one-step transitions from the unaffected region $S_{\mathrm{origin}}$ into the localized repair region $S_{\mathrm{new}}$, while $\Gamma_{\mathrm{out}}^{-}$ and $\Gamma_{\mathrm{out}}^{+}$ describe one-step transitions from $S_{\mathrm{new}}$ back to $S_{\mathrm{origin}}$. The restriction to $\mathcal{H}(\Omega_{\hat{M}}(S_0))$ ensures that incoming transitions originate from states reachable from $S_0$ under the updated system $\hat{M}$. In particular, the entry set of the localized repair region is exactly $S_{0}^{\mathrm{new}}=\Gamma_{\mathrm{in}}^{+}$. For each subsequently introduced unsafe region, the same definitions are instantiated using the corresponding current $S_{\mathrm{origin}}$, $S_{\mathrm{new}}$, and updated system $\widehat{M}$; the sequential construction is detailed in Appendix \ref{app:compositional_safety}.

\begin{definition}[Adaptive Piecewise Barrier Certificates]\label{df:apwbc}
 Given an NNCS $\hat{M}=(S,S_0,A,f,\hat{\pi},R,\hat{S}_u)$, a barrier certificate $B$ w.r.t. $\pi$ and $S_u$ (as defined in~\cref{df:bc}), and a localized barrier certificate $B_{\mathrm{new}}:S\rightarrow \Rset$ w.r.t. $\pi_{\mathrm{new}}$ and $S_u^{\mathrm{new}}$ satisfying
 \begin{align}
&B_{\mathrm{new}}(x) \le 0, && \forall x\in S_{0}^{\mathrm{new}}, \label{eq:bc2-init}\\
&B_{\mathrm{new}}(x) > 0, && \forall x\in S_u^{\mathrm{new}}, \label{eq:bc2-sep}\\
&B_{\mathrm{new}}\!\bigl(f(x,\pi_{\mathrm{new}}(x))\bigr)-\lambda' B_{\mathrm{new}}(x) \le 0,
&& \forall x\in S_{\mathrm{new}}\setminus S_u^{\mathrm{new}}. \label{eq:bc2-inv}
\end{align}
An adaptive piecewise barrier certificate for $\hat{M}$ is a function $V:S\to\mathbb{R}$ such that 
\begin{align*}
 V(x)=
\begin{cases}
B(x), & x\in S_{\mathrm{origin}},\\
\alpha B_{\mathrm{new}}(x), & x\in S_{\mathrm{new}},
\end{cases}
\qquad
\hat{\lambda}(x)=
\begin{cases}
\lambda, & x\in S_{\mathrm{origin}},\\
\lambda', & x\in S_{\mathrm{new}},
\end{cases}   
\end{align*}
with $\alpha >0$ and $\hat{\lambda}(x)\in[0,1]$ for all $x\in S$.
 
\end{definition}

Here, $\alpha B_{\mathrm{new}}$ is also a valid localized barrier certificate on $S_{\mathrm{new}}$ (see more details in Appendix \ref{app:proof_thm2}).
The following theorem gives sufficient conditions under which adaptive piecewise barrier certificates can ensure deployment-time system safety.

\begin{theorem}[Deployment-time Safety by adaptive piecewise barrier certificates]\label{theorem:apwbc}
Consider an NNCS $\hat{M}=(S,S_0,A,f,\hat{\pi},R,\hat{S}_u)$. If there exists an adaptive piecewise barrier certificate $V$ for $\hat{M}$ as defined in~\cref{df:apwbc}, and the switching interface sets satisfy
\begin{equation}\label{eq:interface_set}
\alpha U_b\!\left(B_{\mathrm{new}}(\Gamma_{\mathrm{in}}^+)\right)-
\lambda L_b\!\left(B(\Gamma_{\mathrm{in}}^-)\right)
\le 0,\quad \text{and} \quad
U_b\!\left(B(\Gamma_{\mathrm{out}}^+)\right)- \alpha
\lambda' L_b\!\left(B_{\mathrm{new}}(\Gamma_{\mathrm{out}}^-)\right)
\le 0,
\end{equation}
then the system $\hat{M}$ is safe w.r.t. $\hat{S}_u$ , where $U_b(\cdot)$ and $L_b(\cdot)$ denote the upper and lower bounds of a function over sets, respectively.
\end{theorem}

\cref{theorem:apwbc} shows that deployment-time safety can be reduced to synthesizing a localized barrier certificate $B_{\mathrm{new}}$ on \(S_{\mathrm{new}}\) and verifying its compatibility with the original barrier certificate over the switching interface sets (see more details in Appendix \ref{app:proof_thm2}) . In the next section, we will show how to find $B_{\mathrm{new}}$ at deployment time.

\section{Repair Framework for Safety Restoration}\label{sec:solutions}
\subsection{Overview of the Repair Framework}
\label{sec:framework_overview}

In this section, we introduce our repair framework in~\cref{fig:repair_framework} for safety restoration, which provides formal safety guarantees after system deployment. 
Newly introduced obstacles are identified by perception methods~\cite{elfes2002using,hornung2013octomap} that equipped with our system and taken as input to our framework.

\noindent \textbf{Input.} The input of our framework consists of an NNCS $M$ with a barrier certificate $B$ w.r.t. the control policy $\pi$ and the unsafe set $S_u$, and a newly introduced unsafe set (a.k.a. obstacle) $S^{\mathrm{new}}_u$. 

\noindent \textbf{Step I: Repair Region Construction.} The framework first constructs a localized repair region $S_{\mathrm{new}}$ that covers the newly introduced unsafe set, i.e., $S^{\mathrm{new}}_u\subseteq S_{\mathrm{new}}$ (see details in~\cref{sec:BC transfer}). Then it divides the whole state space $S$ into two disjoint sets $S_{\mathrm{origin}}$ and $S_{\mathrm{new}}$ (see~\cref{sec:Theoretical}). 

\noindent \textbf{Step II: Transfer-based Repair.}
The framework focuses on the repair region $S_{\mathrm{new}}$ and aims to reuse the original policy and barrier certificate whenever possible. This leads to three progressive transfer-based repair stages (see~\cref{sec:BC transfer}).
The goal of Step II is to obtain a policy $\pi_{\mathrm{new}}$ and a valid localized barrier certificate on $S_{\mathrm{new}}$ satisfying~\cref{eq:bc2-init,eq:bc2-sep,eq:bc2-inv}.

\noindent \textbf{Step III: Verification.} After Step II, the framework obtains a localized BC $B_{\mathrm{new}}$ w.r.t. $\pi_{\mathrm{new}}$ and $S_u^{\mathrm{new}}$. Combined with the original BC $B$ w.r.t. $\pi$ and $S_u$, it induces an adaptive piecewise barrier certificate $V$ as defined in~\cref{df:apwbc}. Next, the framework computes the switching interface sets via reachability analysis~\cite{wang2023polar} (see~\cref{set1,set2,set3,set4}) and verifies \cref{eq:interface_set} in \cref{theorem:apwbc}. The computation of upper and lower bounds $U_b(\cdot), L_b(\cdot)$ of BCs over sets is provided in Appendix \ref{app:interface bound}.

\noindent \textbf{Output.} If the verification succeeds, the framework outputs an adaptive piecewise barrier certificate $V$ satisfying the precondition of~\cref{theorem:apwbc}, and thus provides safety guarantee at deployment time.

When multiple newly introduced obstacles are present, the framework processes them sequentially, applying the same localized repair procedure above (See details in Appendix \ref{app:compositional_safety}).

\subsection{Transfer-Based Repair of Localized Barrier Certificates}\label{sec:BC transfer}

\noindent\textbf{Repair Region Construction.}
Given a newly introduced unsafe set $S_u^{\mathrm{new}}$, we construct a localized repair region $S_{\mathrm{new}}$ such that $S_u^{\mathrm{new}}\subseteq S_{\mathrm{new}}\subseteq S$.
Specifically, $S_{\mathrm{new}}$ is defined as a box-shaped expansion of $S_u^{\mathrm{new}}$, i.e., $S_{\mathrm{new}} = S_u^{\mathrm{new}} \oplus \mathcal{N}$, where $\mathcal{N}\subseteq \mathbb{R}^n$ is the repair margin and $\oplus$ denotes the Minkowski sum~\cite{lavalle2006planning, lozano1983spatial} (see more details in Appendix \ref{app:box-shaped-repair-region}).
For example, in a one-dimensional state space, if $S_u^{\mathrm{new}}=[2,3]$ and $\mathcal{N}=[-1,1]$, then $S_{\mathrm{new}}=[2,3]\oplus[-1,1]=[1,4]$.
This construction is commonly used in the literature such as robot motion planning and obstacle avoidance~\cite{best2016real,lien2008hybrid}, and our framework is not restricted to this construction as long as $S_u^{\mathrm{new}}\subseteq S_{\mathrm{new}}\subseteq S$ holds.

\noindent\textbf{Transfer-based Repair.}
Once the repair region \(S_{\mathrm{new}}\) is constructed, we seek a localized BC \(B_{\mathrm{new}}\) satisfying \cref{eq:bc2-init,eq:bc2-sep,eq:bc2-inv} on this region. 
Here we prioritize transfer-based repair over localized retraining by reusing the structure of previously learned certificates, as states in the same state space share similar characteristics and such reuse enables fast adaption of a valid BC. 
Inspired by DNN repair~\cite{goldberger2020minimal,sotoudeh2021provable}, which reduces repair to a linear program, we adopt the last-layer repair strategy. 
Without loss of generality, we consider a transfer template $(S^i,\pi^i,B_i)$, where $S^i$ is a previously certified region. Our formulation is independent of the template-selection strategy.
Let the repair region be $S^j$ controlled by policy $\pi^j$ and its unsafe set is $S^j_u\subseteq S^j$.
We initially set $\pi^j=\pi^i$ and aim to transfer a valid BC $B^j$ from $B^i$.

\noindent\textbf{Template Reuse.}
Let the template certificate be $B_i(x)=w_i^\top\phi(x)$ for $x\in S^i$, where $\phi(\cdot)$ is the feature map defined by the preceding layers and $w_i$ is the last-layer parameter. 
To reuse the feature map on $S^{j}$, we introduce a region-alignment map $T_{ji}:S^{j}\to S^{i}$, which maps states in $S^j$ to states in $S^i$. 
Since both regions are axis-aligned boxes, $T_{ji}$ can be an affine transformation that enables cross-region feature reuse and thus certificate transfer between regions (see an example in~\cref{fig:tij}). Hence, $B_j(x)=w_j^\top \phi(T_{ji}(x))$ for $x\in S^j$, and $w_j$ is an unknown vector to be calculated.

Given the structure of $B_j$, the repair region $S^j$ containing the unsafe set $S_u^j$, and the entry set $S_{0}^j$ (refer to~\cref{set2}). Let $\pi^j=\pi^i_\theta$ ($\theta$ is network parameter), the transferred BC $B_j$ with the unknown vector $w_j$ should satisfy~\cref{eq:bc2-init,eq:bc2-sep,eq:bc2-inv}, which can be written as the following constraints:
\begin{align}
g_{\mathrm{init}}(x;w_j)
&:= w_j^\top \phi(T_{ji}(x)) \le 0,
&& \forall x\in S_{0}^j,
\label{eq:continuous_init}\\
g_{\mathrm{unsafe}}(x;w_j)
&:= -w_j^\top \phi(T_{ji}(x)) \le -\beta,
&& \forall x\in S_{u}^{j},
\label{eq:continuous_unsafe}\\
g_{\mathrm{dyn}}(x;w_j,\theta)
&:= w_j^\top\!\left(
\phi(T_{ji}(f(x,\pi^j(x;\theta))))-\lambda'\phi(T_{ji}(x))
\right) \le 0,
&& \forall x\in S^j\setminus S_u^j,
\label{eq:continuous_dyn}
\end{align}
where \(\beta>0\) is a prescribed safety margin for strict separation from the newly introduced unsafe set.

Existing DNN repair methods address continuous-domain constraints through iterative counterexample-guided repair and verification~\cite{boetius2023robust}.
In contrast, we reduce the continuous BC constraints (\cref{eq:continuous_init,eq:continuous_unsafe,eq:continuous_dyn}) to finite grid constraints with Lipschitz-based margins, enabling a single localized repair step without repeated counterexample generation. To address this, we discretize the state space with granularity $\eta>0$, obtaining finite representative sets while leveraging Lipschitz continuity to ensure validity over the continuous domain.
Specifically, for each state $x$ in a continuous set, there exists a representative $\hat{x}$ such that $\|x-\hat{x}\|_1 \le \eta$. This yields finite discretized sets $\mathcal{D}_{0}^{j}\subseteq S_{0}^j$, $\mathcal{D}_{u}^{j}\subseteq S_{u}^{j}$ and $\mathcal{D}_{\mathrm{dyn}}^{j}\subseteq S^j$.

\begin{theorem}[Continuous validity from discretized transfer constraints]\label{thm:LipBC}
If a function $B_j$ satisfies that 
\begin{align}
g_{\mathrm{init}}(\hat{x};w_j)
&\le -\delta_{\mathrm{init}},
&& \forall \hat{x}\in \mathcal{D}_{0}^{j},
\label{eq:maindisc_init}\\
g_{\mathrm{unsafe}}(\hat{x};w_j)
&\le -(\delta_{\mathrm{unsafe}}+\beta),
&& \forall \hat{x}\in \mathcal{D}_{u}^{j},
\label{eq:maindisc_unsafe}\\
g_{\mathrm{dyn}}(\hat{x};w_j,\theta)
&\le -\delta_{\mathrm{dyn}},
&& \forall \hat{x}\in \mathcal{D}_{\mathrm{dyn}}^{j},
\label{eq:maindisc_dyn}
\end{align}
where constants
$\delta_{\mathrm{init}}$, \(\delta_{\mathrm{unsafe}}\), and \(\delta_{\mathrm{dyn}}\) 
are derived from the granularity \(\eta\) and the relevant Lipschitz constants of the system, then $B_j$ is a valid transferred BC on $S^j$ that satisfies the continuous constraints in~\cref{eq:continuous_init,eq:continuous_unsafe,eq:continuous_dyn}.   
\end{theorem}

\cref{thm:LipBC} ensures that discretizing the continuous constraints still yields valid BCs over the entire state space (see more details in Appendix \ref{app:proof_thm3}) . By setting $S^j=S_{\mathrm{new}}$ and applying \cref{thm:LipBC}, we can obtain a valid localized BC as defined in~\cref{df:apwbc}. As the constraints in \cref{eq:maindisc_init,eq:maindisc_unsafe,eq:maindisc_dyn} are linear in $w_j$ under fixed policy parameter $\theta$, they can be stacked as a \textbf{linear system} $A(\theta)w_j \le -b$ (see Appendix \ref{app:sec5} for details). To obtain \(w_j\), we introduce a nonnegative slack variable \(\xi\), require $w_j$ to be bounded by some constant $W>0$, i.e., $\|w_j\|_1\le W$ and solve the following linear program:
\begin{equation} \label{eq:mainstage1_target}
\begin{aligned}
\mathbf{(P_{\mathrm{I}})}\qquad
\min_{w_j,\xi,u_j}\quad & \mathbf{1}^{\top}\xi \\
\mathrm{s.t.}\quad
& A(\theta)w_j \le -b+\xi, -u_j\le w_j\le u_j, \mathbf{1}^{\top}u_j\le W,\xi\ge 0, u_j\ge0 .
\end{aligned}
\end{equation}

\noindent \textbf{Stage I: Last-layer certificate repair under the template policy.} 
Stage I solves the linear program \(\mathbf{(P_{\mathrm{I}})}\) in \cref{eq:mainstage1_target} with the policy parameter $\theta$ fixed from the template policy $\pi^i_\theta$. Under fixed $\theta$, the constraints are linear in the decision variables \(w_j\), \(\xi\) and \(u_j\). Let \((w_{j,\mathrm{I}}^\star,\xi_{\mathrm{I}}^\star, u_{j,\mathrm{I}}^\star)\) be an optimal solution of \(\mathbf{(P_{\mathrm{I}})}\), and define the aggregate slack $\varepsilon_{\mathrm{I}}^\star := \mathbf{1}^{\top}\xi_{\mathrm{I}}^\star$, which measures the residual violation of the linear system $A(\theta)w_j \le -b$.

\begin{theorem}[Feasibility for Stage I ]
Assume that \(\mathbf{(P_{\mathrm{I}})}\) attains an optimal solution. Then $\varepsilon_{\mathrm{I}}^\star=0$ if and only if there exists a last-layer parameter \(w_j\) satisfying $A(\theta)w_j\le -b$. Consequently, Stage~I succeeds and yields a valid localized barrier certificate as defined in~\cref{df:apwbc}.
\label{thm:mainstage1_success}
\end{theorem}

If $\varepsilon_{\mathrm{I}}^\star>0$, it means that Stage I cannot yield a valid localized barrier certificate. The framework therefore proceeds to Stage II, where the policy parameter is allowed to be updated.
The proof is in Appendix \ref{app:prove_thm4}.

\noindent \textbf{Stage II: Joint policy adaptation and last-layer certificate repair.}
Stage~II adapts the local policy parameter together with the last layer of the transferred barrier certificate, which aims to find a pair \((\theta,w_j)\) such that $A(\theta)w_j \le -b$ holds.
Let $(w_{j,\mathrm{II}}^\star(\theta),\xi_{\mathrm{II}}^\star(\theta),u_{j,\mathrm{II}}^\star(\theta) )$
an optimal solution of the lower-level problem
for the current $\theta$ and the corresponding optimal aggregate slack be $\varepsilon_{\mathrm{II}}(\theta):=\mathbf{1}^{\top}\xi_{\mathrm{II}}^\star(\theta)$. Stage~II solves the following bilevel optimization problem:
\begin{equation}\label{eq:mainbilevel}
\begin{aligned}
\mathbf{(P_{\mathrm{II}})}\qquad
\min_{\theta}\quad & \varepsilon_{\mathrm{II}}(\theta) \\
\mathrm{s.t.}\quad
& (w_{j,\mathrm{II}}^\star(\theta),\xi_{\mathrm{II}}^\star(\theta),u_{j,\mathrm{II}}^\star(\theta) )
\in
\arg\min_{w_j,\xi,u_j}\ \mathbf{1}^{\top}\xi \\
& \hspace{0.2cm}
\mathrm{s.t.}\quad
A(\theta)w_j \le -b+\xi, -u_j\le w_j\le u_j,\mathbf{1}^{\top}u_j\le W,\xi\ge 0, u_j\ge0.
\end{aligned}
\end{equation}

The upper-level problem searches for a policy parameter \(\theta\) that reduces the optimal aggregate slack \(\varepsilon_{\mathrm{II}}(\theta)\), while the lower-level problem computes the optimal last-layer repair \(w_j\) for the current \(\theta\).

\begin{theorem}[Feasibility for Stage II]
\label{thm:mainstage2_success}
Assume that \(\mathbf{(P_{\mathrm{II}})}\) attains an optimal solution. Then $\varepsilon_{\mathrm{II}}(\theta)=0$
if and only if there exists a tuple \((w_j,\theta)\) such that $A(\theta)w_j\le -b $.
Consequently, Stage~II succeeds and yields a valid localized barrier certificate as defined in~\cref{df:apwbc}.
\end{theorem}

In practice, Stage~II solves \(\mathbf{(P_{\mathrm{II}})}\) iteratively. Starting from an initial policy parameter \(\theta\), it solves the lower-level problem \(\mathbf{(P_{\mathrm{I}}(\theta))}\) to obtain the aggregate slack \(\varepsilon_{\mathrm{II}}(\theta)\). If \(\varepsilon_{\mathrm{II}}(\theta)=0\), then the current \(\theta\) and the recovered \(w_{j,\mathrm{II}}^\star(\theta)\) satisfy $A(\theta)w_j \le -b$, and Stage~II succeeds. Otherwise, the nonzero slack indicates that the barrier constraints remain violated under the current policy. In this case, we use the slack as a repair loss to update \(\theta\) via backpropagation~\cite{rumelhart1986learning},\footnote{The lower-level LP constraints in~\cref{eq:mainbilevel} are recalculated after each policy update.} and repeat the process until feasibility is achieved; otherwise, the framework proceeds to Stage~III once the computational budget is exhausted. The proof is in Appendix \ref{app:prove_thm5}.

\noindent \textbf{Stage III: Localized retraining both the policy and the barrier certificate.}
Stage~III performs localized retraining on \(S_{\mathrm{new}}^j\) to synthesize both the policy and the barrier certificate (satisfying~\cref{eq:bc2-init,eq:bc2-sep,eq:bc2-inv}) tailored to the repair region.
When Stage~III succeeds, it outputs a valid localized BC as defined in~\cref{df:apwbc}.

Once a localized BC is obtained, Step III (\cref{sec:framework_overview}) is triggered to ensure deployment-time safety.

\section{Evaluation}\label{sec:evaluation}

We evaluate the proposed framework~(\cref{sec:framework_overview}) from four aspects: (i) the end-to-end repair under multiple obstacles, assessing whether it restores global safety efficiently and outperforms global retraining in preserving nominal performance; (ii) the effect of the repair region size $S_{\mathrm{new}}$ on computational cost, examining the trade-off introduced by region construction; (iii) the impact of template selection on repair efficiency, evaluating robustness across different transfer templates; and (iv) the comparison with the shielding method, showcasing the cumulative runtime overhead and task performance under persistent environmental changes.
We measure overhead from the initial detection of each newly introduced obstacle, excluding the detection cost shared by our method and shielding to ensure a fair comparison.

\noindent\textbf{Experimental Setup.}
The proposed framework is tested on four benchmarks: B1, B2~\cite{gallestey2019lecture}, Mountain Car Continuous (MC), and Pendulum (PD)~\cite{brockman2016openai}. For each benchmark, we introduce new obstacles that are randomly placed, and construct each repair region by box-shaped expansion (see Appendix ~\ref{app:box-shaped-repair-region}). The DRL algorithms used for NNCSs and NBCs are DDPG~\cite{lillicrap2020continuous} for B1 and B2, PPO~\cite{schulman2017proximal} for MC, and SAC~\cite{haarnoja2018soft} for PD.
Experiments are implemented in Python~3.8.0 and PyTorch~2.3.1, and run on a workstation with a 32-core AMD Ryzen Threadripper CPU, 128GB RAM, and a 24GB GPU.

\subsection{End-to-End Repair under Multiple Obstacles}\label{sec:main expriment}

\begin{figure}[!htbp]
    \centering
    \begin{minipage}[t]{0.43\linewidth}
        \centering
        \includegraphics[width=\linewidth,trim={10pt 9pt 10pt 7pt},
  clip]{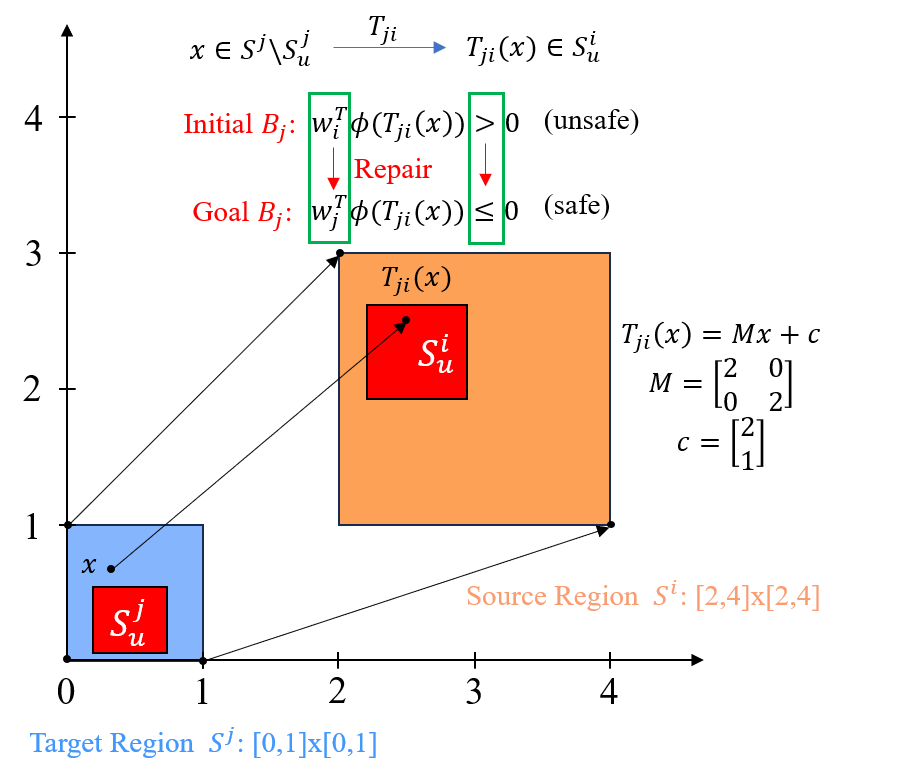}
        \caption{Illustration of $T_{ji}$ and template reuse}
        \label{fig:tij}
    \end{minipage}
    \hfill
    \begin{minipage}[t]{0.55\linewidth}
        \centering
        \includegraphics[width=\linewidth,trim={10pt 45pt 10pt 8pt},
  clip]{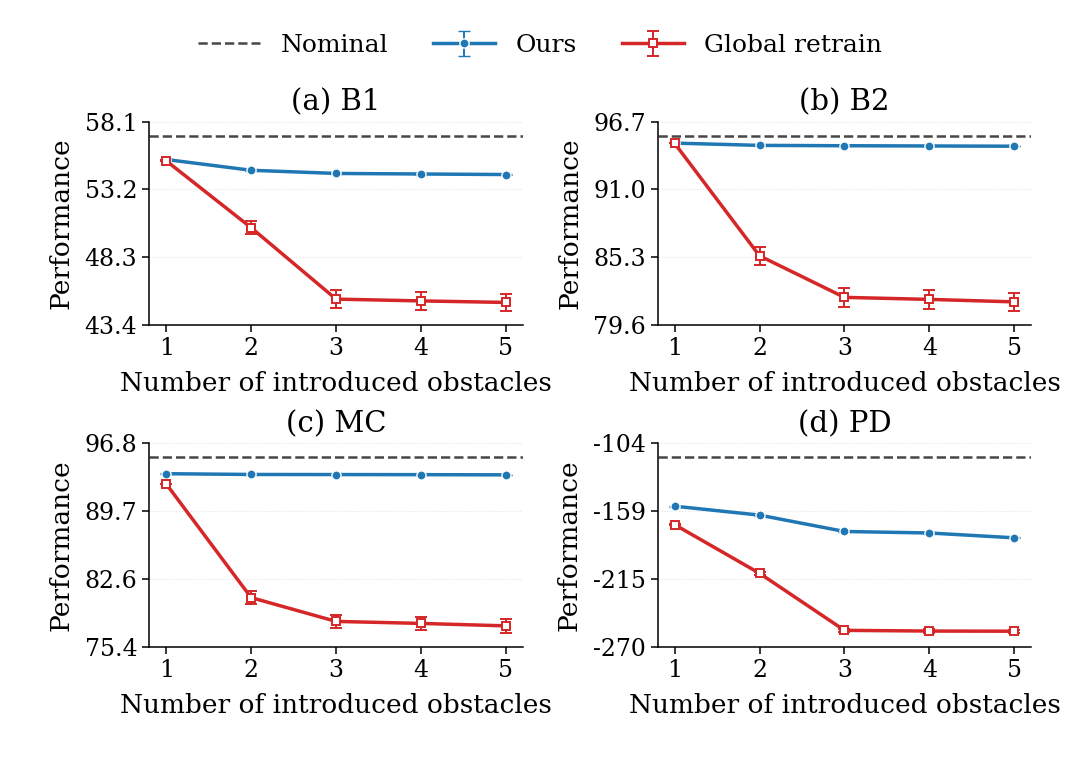}
        \caption{Performance (average cumulative reward) under increasing numbers of newly introduced obstacles on all benchmarks. The error bar applies standard error.}
    \label{fig:reward_multi_obstacles}
    \end{minipage}
\end{figure}

We evaluate our repair framework under multiple newly introduced obstacles, and use $O_i$ to denote the \textit{i}-th obstacle.
By default, we use the previous-template strategy: the original barrier certificate is used for the first obstacle, and each repaired localized certificate is reused as the source template for the next obstacle. 
\cref{tab:main_multi_obstacle} reports a representative subset of the runtime results, with complete results of the same claims in Appendix \ref{app:detailed_results}. 
We also evaluate system performance by the average cumulative reward over $1000$ simulations, as shown in \cref{fig:reward_multi_obstacles}.

\cref{tab:main_multi_obstacle} shows that global retraining becomes infeasible when multiple obstacles are introduced: it times out from \(O_2\) on B1, B2, and MC, and already fails on \(O_1\) for PD. In contrast, our repair framework restores a valid loalized barrier certificate for all obstacles and achieves \textbf{17.3$\times$} faster on average on the instances where retraining succeeds. After the first repair, most subsequent obstacles are handled by Stage I alone, with average runtime of 13.9s, 10.9s, 28.85s and 84.6s on B1, B2, MC and PD, respectively.
For MC-O1 and PD-O1 where Stage~II times out, the framework invokes Stage~III (localized retraining) as a warm-up step. After this warm-up step, the remaining obstacles are handled efficiently by Stage~I alone. 

For performance, rewards are computed using the original benchmark reward function with an additional $-50$ collision penalty. As shown in \cref{fig:reward_multi_obstacles}, after five obstacles, our method incurs only 16.2\% average degradation, whereas global retraining often times out and yields no certified safe policy, causing collisions and larger performance loss. 
The nominal policy serves as a performance reference, but its
certificate only guarantees safety for the original unsafe set
$S_u$, requiring repair and verification to restore the safety
guarantee under $\widehat{S}_u$.
These results show that our repair framework restores deployment-time safety while preserving nominal performance.

\begin{table*}[!htbp]
\centering

\caption{Runtime (in seconds) comparison under multiple obstacles across benchmarks.}
\label{tab:main_multi_obstacle}
\resizebox{0.8\textwidth}{!}{
\begin{tabular}{c c c c c c c c c}
\toprule
\multirow{2}{*}{Benchmark}
& \multirow{2}{*}{Original BC}
& \multicolumn{2}{c}{Baseline}
& \multicolumn{5}{c}{Ours} \\
\cmidrule(lr){3-4} \cmidrule(lr){5-9}
&
& Obstacle ID
& Global retrain
& Stage I
& Stage II
& Stage III
& Verification
& Total time \\
\midrule
\multirow{3}{*}{B1}
& \multirow{3}{*}{616.6}
& $O_1$ & 1516.3 & 10.6 \ding{55}  & 55.3 \checkmark & 0 & 2.0 & 67.9  \\
&
& $O_2$ & F      & 10.3 \checkmark & 0               & 0 & 3.2 & 13.5  \\
&
& $O_3$ & F      & 11.2 \checkmark & 0               & 0 & 3.1 & 14.3  \\
\midrule
\multirow{3}{*}{B2}
& \multirow{3}{*}{685.3}
& $O_1$ & 1309.5 & 8.2 \ding{55}   & 40.7 \checkmark & 0 & 1.2 & 50.1  \\
&
& $O_2$ & F      & 8.1 \ding{55}   & 44.8 \checkmark & 0 & 1.5 & 54.4  \\
&
& $O_3$ & F      & 8.9 \checkmark  & 0               & 0 & 2.0 & 10.9  \\
\midrule
\multirow{3}{*}{MC}
& \multirow{3}{*}{796.3}
& $O_1$ & 1114.0 & 11.5 \ding{55}  & 120 \ding{55}& 173.4 \checkmark & 16.7 & 321.6 \\
&
& $O_2$ & F      & 12.2 \checkmark & 0            & 0                & 16.7 & 28.9  \\
&
& $O_3$ & F      & 11.9 \checkmark & 0            & 0                & 16.9 & 28.8  \\
\midrule
\multirow{3}{*}{PD}
& \multirow{3}{*}{1259.5}
& $O_1$ & F & 14.4 \ding{55}  & 240 \ding{55} & 637.1 \checkmark & 40.8 & 932.3 \\
&
& $O_2$ & F & 14.8 \checkmark & 0             & 0     & 68.1 & 82.9  \\
&
& $O_3$ & F & 14.5 \checkmark & 0             & 0     & 71.8 & 86.3  \\
\bottomrule
\end{tabular}
}
\vspace{0.5ex}
\footnotesize

\noindent\textbf{Remark.}
\textbf{Original BC}: time for synthesizing the original barrier certificate for $S_0$ and $S_u$;
\textbf{Verification}: time for verifying \cref{theorem:apwbc};
\textbf{Total time}: end-to-end runtime of our repair framework;  ``F'' means retrain failed within 7200s; $\checkmark$: Stage succeeds; \ding{55}: Stage fails.

\end{table*}

\subsection{Effect of Repair Region Size}\label{subsec:size_sensitivity}

We investigate how the size of the repair region \(S_{\mathrm{new}}\) affects repair feasibility and runtime. For each benchmark, we fix the third obstacle $O_3$  and use the localized repaired certificate for $O_2$ as the source template, so that only the size of \(S_{\mathrm{new}}\) varies. Starting from the default neighborhood \(\mathcal{N}\), we scale it by a factor \(\rho\) and construct \(S_{\mathrm{new}}(\rho)=S_u^{\mathrm{new}}\oplus(\rho \mathcal{N})\), with \(\rho\in\{0.6,0.8,1.0,1.2,1.4\}\). ~\cref{tab:size_sensitivity} reports representative results, while full results with the same claim are provided in Appendix \ref{app:detailed_results}.

\cref{tab:size_sensitivity} shows a clear trade-off in choosing \(S_{\mathrm{new}}\). Small regions lead to infeasibility (\(0.6\times\) fails on both MC and PD), while moderate sizes allow efficient repair via Stage~I alone (8.2--11.9s on MC and 10.5--14.5s on PD for \(0.8\times\) and \(1.0\times\), respectively). Large regions increase cost and require Stage~II (95.0s on MC and 159.8s on PD at \(1.2\times\), respectively), and may incur Stage~III (\(1.4\times\) with high runtime). This suggests choosing \(S_{\mathrm{new}}\) large enough for feasibility but small enough to limit cost.
In practice, doubling the obstacle size provides a reasonable starting point for constructing \(S_{\mathrm{new}}\).

\begin{table*}[!htbp]
\centering
\caption{Runtime (in seconds) under different scaling factors of repair regions.}
\label{tab:size_sensitivity}
\resizebox{0.55\textwidth}{!}{
\begin{tabular}{c c c c c c}
\toprule
Benchmark & Scale of $S_{\mathrm{new}}$ & Stage I & Stage II & Total time of Stages~I,II & Stage III \\
\midrule
\multirow{5}{*}{MC}
& $0.6\times$ & 5.1 \ding{55} & F    & F     & F \\
& $0.8\times$ & 8.2 \checkmark & 0    & 8.2  & 0 \\
& $1.0\times$ & 11.9 \checkmark & 0    & 11.9 & 0 \\
& $1.2\times$ & 17.7 \ding{55} & 77.3 \checkmark & 95.0 & 0 \\
& $1.4\times$ & 22.3 \ding{55} & F   & F & 402.5 \checkmark \\
\midrule
\multirow{5}{*}{PD}
& $0.6\times$ & 6.4 \ding{55} & F  & F  & F \\
& $0.8\times$ & 10.5 \checkmark & 0     & 10.5 & 0 \\
& $1.0\times$ & 14.5 \checkmark & 0     & 14.5 & 0 \\
& $1.2\times$ & 22.0 \ding{55} & 137.8 \checkmark &159.8 & 0 \\
& $1.4\times$ & 28.3 \ding{55} & F  & F &  862.6 \checkmark\\
\bottomrule
\end{tabular}
}
\vspace{0.5em}
\footnotesize

\textbf{Remark.}
Stage~III denotes local retraining in the scaled repair region.
Two stages total denotes the total running time of Stage~I and Stage~II for the current scaled region.
``F'' denotes failure due to timeout in the transfer stage.
\end{table*}

\subsection{Impact of Template Selection}

We study how template selection affects transfer-based repair efficiency via three strategies: (i) using the original global certificate, (ii) using the first repaired localized certificate when available, and (iii) using the previously repaired certificate, as the source template. For the first obstacle, where no repaired template exists, the previous-template strategy is initialized with the original global certificate. ~\cref{tab:template-selection} reports representative results on MC and PD, with full results in Appendix~\ref{app:detailed_results}.

\cref{tab:template-selection} shows that template selection is critical on MC and PD. Using Strategy (i) leads to Stages~I,II timeout across all cases, whereas using Strategy (ii) or (iii) resolves \(O_2\)$\sim$\(O_3\) via Stage~I alone. Strategy (iii) thus provides a simple default for sequential repair, avoiding repeated reliance on the original global certificate, as the previous repaired templates better capture the updated environment.

\begin{table}[!htbp]
\centering
\caption{Running time (in seconds) comparison of template selection across benchmarks.}
\label{tab:template-selection}
\resizebox{0.75\textwidth}{!}{
\begin{tabular}{c c ccc ccc ccc}
\toprule
\multirow{2}{*}{Benchmark}
& \multirow{2}{*}{Obstacle ID}
& \multicolumn{3}{c}{Original global BC }
& \multicolumn{3}{c}{First repaired local BC}
& \multicolumn{3}{c}{Previous template} \\
\cmidrule(lr){3-5}
\cmidrule(lr){6-8}
\cmidrule(lr){9-11}
& 
& Stage I & Stage II & Total time
& Stage I & Stage II & Total time
& Stage I & Stage II & Total time \\

\midrule
\multirow{3}{*}{MC}
& O1
&11.5& F & F
& -- & -- & --
&11.5& F & F   \\
& O2
& 11.6 & F & F
& 12.2 & 0 & 12.2
& 12.2 & 0 & 12.2 \\
& O3
& 12.1 & F & F
& 11.8 & 0 & 11.8
& 11.9 & 0 & 11.9 \\

\midrule
\multirow{3}{*}{PD}
& O1
& 14.4 & F & F
& -- & -- & --
& 14.4 & F & F \\
& O2
& 14.0 & F & F
& 14.8 & 0 & 14.8
& 14.8 & 0 & 14.8 \\
& O3
& 14.2 & F & F
& 14.0 & 0 & 14.0
& 14.5 & 0 & 14.5 \\

\bottomrule
\end{tabular}
}
\vspace{0.5em}
\begin{flushleft}
\footnotesize
\textbf{Remark.}
"F" denotes failure due to timeout in the transfer stage. $-$ means no repaired BC available and thus skip repair.
\end{flushleft}
\end{table}

\begin{table*}[!htbp]
\centering
\caption{Comparison between our method and shielding under multiple obstacles across benchmarks.}
\label{tab:shield_multi_obstacle}
\resizebox{0.8\textwidth}{!}{%
\begin{tabular}{c c c c c c c c c}
\toprule
\multirow{2}{*}{Benchmark}
& \multirow{2}{*}{\# Obs.}
& \multicolumn{3}{c}{Ours}
& \multicolumn{4}{c}{Shielding} \\
\cmidrule(lr){3-5} \cmidrule(lr){6-9}
&
& Total runtime
& Avg. reward
& Success rate (\%)
& \# Shield Interventions
& Total runtime
& Avg. reward
& Success rate (\%) \\
\midrule
\multirow{3}{*}{B1}
& $O_1$ & 67.9 & 57.10 & 100 &622 & 269.8 & 56.14 & 100 \\
& $O_2$ & 81.4 & 55.37 & 100 &972 & 413.6 & 53.96 & 100 \\
& $O_3$ & 95.7 & 54.59 & 100 &1354 & 573.2 & 51.30 & 97  \\
\midrule
\multirow{3}{*}{B2}
& $O_1$ & 50.1  & 95.11 & 100 & 241& 102.0 & 94.83 & 100 \\
& $O_2$ & 104.5 & 94.89 & 100 &270 & 116.2 & 94.12 & 100 \\
& $O_3$ & 115.4 & 94.70 & 100 &389 & 158.6 & 93.05 & 99  \\
\bottomrule
\end{tabular}
}
\vspace{0.5ex}
\footnotesize

\noindent\textbf{Remark.}
\textbf{Total runtime}: total runtime in seconds over 200 executions;
\textbf{Avg. reward}: the average cumulative reward;
\textbf{ \# Shield Interventions}: number of intervention times by shielding.

\end{table*}

\subsection{Comparison with Shielding Method}

We compare our method with shielding~\cite{banerjee2024dynamic}
over 200 executions per benchmark and obstacle configuration,
evaluating empirical success rate, average cumulative reward,
and total runtime.
The runtime includes one-time repair costs for our method and
all recurring safety checks and interventions for shielding.
\Cref{tab:shield_multi_obstacle} presents representative results
on B1 and B2, with full results in Appendix~\ref{app:detailed_results}.

Our method achieves 100\% empirical success, higher average
rewards, and lower cumulative runtime across all configurations.
In contrast, shielding achieves 97\% on B1-$O_3$ and 99\%
on B2-$O_3$. On B1-$O_3$, our method requires 95.7\,s
versus 573.2\,s for shielding, with average rewards of
54.59 versus 51.30. Shielding incurs 241--1,354 interventions
across configurations, accumulating overhead upon repeated
encounters with persistent obstacles. Our method instead
repairs each new obstacle once and reuses the certified
system across subsequent executions.

\section{Conclusion}\label{sec:conclusion}

In this paper, we propose an adaptive piecewise barrier certificate framework for deployment-time safety restoration of NNCSs in response to newly introduced unsafe regions. By restricting adaptation to the affected regions while preserving the original policy and certificate elsewhere, the framework enables efficient localized repair with global safety guarantees. It avoids retraining from scratch and largely preserves nominal policy performance.
A current limitation is that transfer-based certificate repair is restricted to the last layer of the neural BC, which may limit its flexibility in more complex scenarios. In future work, we plan to extend the repair to multiple layers and further investigate how the locations of newly introduced obstacles affect repair feasibility and computational efficiency.

\begin{ack}
 We thank the anonymous reviewers for their constructive comments and valuable suggestions. This work was partially supported by the National Natural Science Foundation of China (Grant No. 62602204, No. 62372176 and No. 92582108); the CCF-Huawei Populus Grove Fund; Shanghai Trusted Industry Internet Software Collaborative Innovation Center; the Basic Research Program of  Institute of Software, CAS (Grant No. ISCAS-JCMS-202302); the National Research Foundation, Singapore, under the NRF RSS Scheme NRF-RSS2022-009 and the Singapore Global AI VP grant AIVP-2024-002.
\end{ack}
\bibliographystyle{unsrt}
\bibliography{ref}
\newpage
\appendix
\section{Technical Appendices and Supplementary Material}

\subsection{Proof of \cref{theorem:apwbc}}
\label{app:proof_thm2}
In this section, we prove \cref{theorem:apwbc} in the main pages. First, we present a proposition.
\begin{proposition}\label{propo:1}
If $B(x)$ is a barrier certificate for a system $M$, then the $\alpha B(x)$ is also a barrier certificate for system $M$, $\alpha > 0,\alpha \in \mathbb{R}$.
\end{proposition}
\begin{proof}

$B(x)$is a barrier certificate for $M$, so it satisfies the following conditions:
\begin{align*}
    & B(x)\le 0 & \forall x\in S_0 \\
    & B(x)> 0 & \forall x\in S_u \\
    & B(f(x,\pi (x)))- \lambda B(x)\le 0 & \forall x\in S\setminus S_u
\end{align*}

Because $\alpha>0$, so we can get:
\begin{align*}
    & \alpha B(x)\le 0 & \forall x\in S_0 \\
    & \alpha B(x)> 0 & \forall x\in S_u \\
    & \alpha B(f(x,\pi (x)))-\lambda \alpha B(x)\le 0 & \forall x\in S\setminus S_u
\end{align*}

So, $\alpha B(x)$ is also a barrier certificate for system $M$.
\end{proof}

Then, we present the definition of piecewise barrier certificate and its safety theorem.

\begin{definition}[Piecewise Barrier Certificates]\label{df:pwBC}
Given a DNN-controlled system $M=(S,S_0,A,f,\pi,R,S_u)$ and its extension $\hat{M}=(S,S_0,A,f,\hat{\pi},R,\hat{S}_u)$, a \emph{piecewise barrier certificate} is a function $V:S\to\mathbb{R}$ such that
\begin{align}
    &V(x)\le 0, && \forall x\in S_0, \label{eq:condition1}\\
    &V(x)>0, && \forall x\in \hat{S}_u, \label{eq:condition2}\\
    &V\bigl(f(x,\hat{\pi}(x))\bigr)-\hat{\lambda}(x)V(x)\le 0, && \forall x\in S\setminus\hat{S}_u, \label{eq:condition3}
\end{align}
where
\[
V(x)=
\begin{cases}
V_1(x), & x\in S_{\mathrm{origin}},\\
V_2(x), & x\in S_{\mathrm{new}},
\end{cases}
\qquad
\hat{\lambda}(x)=
\begin{cases}
\lambda, & x\in S_{\mathrm{origin}},\\
\lambda', & x\in S_{\mathrm{new}},
\end{cases}
\]
with $\hat{\lambda}(x)\in[0,1]$ for all $x\in S$.
\end{definition}

\begin{theorem}[Safety guarantee of a piecewise barrier certificate]\label{thm:piecewise}
If there exists a piecewise barrier certificate for the extended system $\hat{M}$, then any trajectory of $\hat{M}$ under policy $\hat{\pi}$ never enters the unsafe set $\hat{S}_u$.
\end{theorem}

\begin{proof}
Let $\{x_t\}_{t\in\mathbb{N}_0}$ be any trajectory of the extended system $\hat{M}$ under the piecewise policy $\hat{\pi}$, with $x_0\in S_0$. By \cref{eq:condition1}, we have
\[
V(x_0)\le 0 .
\]
We prove by induction that
\[
V(x_t)\le 0,\qquad \forall t\in\mathbb{N}_0 .
\]

Suppose that \(V(x_t)\le 0\) holds for some \(t\in\mathbb{N}_0\). By \cref{eq:condition2}, every state in
\[
\hat{S}_u=S_u\cup S_u^{\mathrm{new}}
\]
has strictly positive \(V\)-value. Therefore, \(V(x_t)\le 0\) implies
\[
x_t\notin \hat{S}_u .
\]
Hence, \(x_t\in S\setminus \hat{S}_u\), and \cref{eq:condition3} applies:
\[
V\bigl(f(x_t,\hat{\pi}(x_t))\bigr)
-
\hat{\lambda}(x_t)V(x_t)
\le 0 .
\]

We now prove that \(V(x_{t+1})\le 0\) by considering the region containing the current state \(x_t\). Since the state space is partitioned into \(S_{\mathrm{origin}}\) and \(S_{\mathrm{new}}\), there are two main cases.

\textbf{Case 1: \(x_t\in S_{\mathrm{origin}}\).}
In this case, the active policy, certificate branch, and decay factor are
\[
\hat{\pi}(x_t)=\pi(x_t),\qquad
V(x_t)=V_1(x_t),\qquad
\hat{\lambda}(x_t)=\lambda .
\]
Thus, \(x_{t+1}=f(x_t,\pi(x_t))\), and \cref{eq:condition3} gives
\[
V(x_{t+1})-\lambda V_1(x_t)\le 0 .
\]
Since \(V_1(x_t)=V(x_t)\le 0\) and \(\lambda\in[0,1]\), we obtain
\[
V(x_{t+1})\le \lambda V_1(x_t)\le 0 .
\]
To make the piecewise structure explicit, we further distinguish the region of \(x_{t+1}\).

If \(x_{t+1}\in S_{\mathrm{origin}}\), then \(V(x_{t+1})=V_1(x_{t+1})\), and hence
\[
V_1(x_{t+1})\le \lambda V_1(x_t)\le 0 .
\]
If \(x_{t+1}\in S_{\mathrm{new}}\), then \(V(x_{t+1})=V_2(x_{t+1})\), and hence
\[
V_2(x_{t+1})\le \lambda V_1(x_t)\le 0 .
\]
Therefore, in both subcases under \(x_t\in S_{\mathrm{origin}}\), we have \(V(x_{t+1})\le 0\).

\textbf{Case 2: \(x_t\in S_{\mathrm{new}}\).}
In this case, the active policy, certificate branch, and decay factor are
\[
\hat{\pi}(x_t)=\pi_{\mathrm{new}}(x_t),\qquad
V(x_t)=V_2(x_t),\qquad
\hat{\lambda}(x_t)=\lambda' .
\]
Thus, \(x_{t+1}=f(x_t,\pi_{\mathrm{new}}(x_t))\), and \cref{eq:condition3} gives
\[
V(x_{t+1})-\lambda' V_2(x_t)\le 0 .
\]
Since \(V_2(x_t)=V(x_t)\le 0\) and \(\lambda'\in[0,1]\), we obtain
\[
V(x_{t+1})\le \lambda' V_2(x_t)\le 0 .
\]
Again, we distinguish the region of \(x_{t+1}\).

If \(x_{t+1}\in S_{\mathrm{new}}\), then \(V(x_{t+1})=V_2(x_{t+1})\), and hence
\[
V_2(x_{t+1})\le \lambda' V_2(x_t)\le 0 .
\]
If \(x_{t+1}\in S_{\mathrm{origin}}\), then \(V(x_{t+1})=V_1(x_{t+1})\), and hence
\[
V_1(x_{t+1})\le \lambda' V_2(x_t)\le 0 .
\]
Therefore, in both subcases under \(x_t\in S_{\mathrm{new}}\), we have \(V(x_{t+1})\le 0\).

Combining the two main cases, we have shown that
\[
V(x_t)\le 0 \quad \Longrightarrow \quad V(x_{t+1})\le 0 .
\]
By induction,
\[
V(x_t)\le 0,\qquad \forall t\in\mathbb{N}_0 .
\]

Finally, by \cref{eq:condition2}, every state in \(\hat{S}_u\) has strictly positive \(V\)-value:
\[
V(x)>0,\qquad \forall x\in \hat{S}_u .
\]
Since every state along the trajectory satisfies \(V(x_t)\le 0\), no trajectory state can belong to \(\hat{S}_u\). Hence,
\[
x_t\notin \hat{S}_u,\qquad \forall t\in\mathbb{N}_0 .
\]
Therefore, every trajectory of \(\hat{M}\) starting from \(S_0\) remains safe.
\end{proof}

Then we can prove \cref{theorem:apwbc}.

\noindent\textbf{Theorem \ref{theorem:apwbc}} (Deployment-time Safety by adaptive piecewise barrier certificates).
Consider an NNCS $\hat{M}=(S,S_0,A,f,\hat{\pi},R,\hat{S}_u)$. If there exists an adaptive piecewise barrier certificate $V$ for $\hat{M}$ as defined in~\cref{df:apwbc}, and the switching interface sets satisfy
\begin{equation}
 \alpha U_b\!\left(B_{\mathrm{new}}(\Gamma_{\mathrm{in}}^+)\right)-
\lambda L_b\!\left(B(\Gamma_{\mathrm{in}}^-)\right)
\le 0,\quad \text{and} \quad
U_b\!\left(B(\Gamma_{\mathrm{out}}^+)\right)-\alpha
\lambda' L_b\!\left(B_{\mathrm{new}}(\Gamma_{\mathrm{out}}^-)\right)
\le 0,  \tag{\ref{eq:interface_set}}
\end{equation}
then the system $\hat{M}$ is safe, where $U_b(\cdot)$ and $L_b(\cdot)$ denote the upper and lower bounds of a function over a set, respectively.

\begin{proof}
We prove that the adaptive piecewise barrier certificate \(V\) satisfies the three conditions in \cref{df:pwBC}. 

First, we verify the \cref{eq:condition1}. Since the admissible initial set belongs to the unaffected region, i.e.,
\(S_0\subseteq S_{\mathrm{origin}}\), and \(B\) is a barrier certificate for the original system on
\(S_{\mathrm{origin}}\), we have
\[
V(x)=B(x)\le 0,\qquad \forall x\in S_0 .
\]
Thus, the \cref{eq:condition1} in \cref{df:pwBC} holds.

Second, we verify \cref{eq:condition2}. For the original unsafe set \(S_u\), since
\(V(x)=B(x)\) on \(S_{\mathrm{origin}}\) and \(B\) separates \(S_u\), we have
\[
V(x)=B(x)>0,\qquad \forall x\in S_u .
\]
For the newly introduced unsafe set \(S_u^{\mathrm{new}}\), since
\(S_u^{\mathrm{new}}\subseteq S_{\mathrm{new}}\), \(V(x)=\alpha B_{\mathrm{new}}(x)\), and
\(B_{\mathrm{new}}\) satisfies \cref{eq:bc2-sep}, we have
\[
V(x)=\alpha B_{\mathrm{new}}(x)>0,\qquad \forall x\in S_u^{\mathrm{new}} .
\]
Therefore,
\[
V(x)>0,\qquad \forall x\in \hat{S}_u=S_u\cup S_u^{\mathrm{new}} .
\]
Thus, the \cref{eq:condition2} in \cref{df:pwBC} holds.

It remains to prove \cref{eq:condition3}. Let
\[
x\in S\setminus \hat{S}_u,
\qquad
x^+=f(x,\hat{\pi}(x)).
\]
We need to show that
\[
V(x^+)-\hat{\lambda}(x)V(x)\le 0 .
\]
Following the piecewise structure of \(V\), \(\hat{\pi}\) and \(\hat{\lambda}\), we divide the proof into two main cases according to the region containing the current state \(x\).

\textbf{Case 1: \(x\in S_{\mathrm{origin}}\).}
In this case,
\[
V(x)=B(x),\qquad
\hat{\pi}(x)=\pi(x),\qquad
\hat{\lambda}(x)=\lambda,
\]
and hence
\[
x^+=f(x,\pi(x)).
\]
We further distinguish two subcases according to the region containing \(x^+\).

\emph{Subcase 1.1: \(x^+\in S_{\mathrm{origin}}\).}
Then \(V(x^+)=B(x^+)\). Since \(B\) is a barrier certificate on \(S_{\mathrm{origin}}\), the \cref{eq:bc-inv} gives
\[
B(f(x,\pi(x)))-\lambda B(x)\le 0 .
\]
Therefore,
\[
V(x^+)-\hat{\lambda}(x)V(x)
=
B(x^+)-\lambda B(x)
\le 0 .
\]

\emph{Subcase 1.2: \(x^+\in S_{\mathrm{new}}\).}
Then the transition crosses from \(S_{\mathrm{origin}}\) to \(S_{\mathrm{new}}\). By the definition of the switching interface sets,
\[
x\in \Gamma_{\mathrm{in}}^-,
\qquad
x^+\in \Gamma_{\mathrm{in}}^+ .
\]
In this case,
\[
V(x)=B(x),\qquad V(x^+)=\alpha B_{\mathrm{new}}(x^+),
\]
so the \cref{eq:condition3} becomes
\[
\alpha B_{\mathrm{new}}(x^+)-\lambda B(x)\le 0 .
\]
Using the upper and lower bounds over the interface sets, we have
\[
\alpha B_{\mathrm{new}}(x^+)
\le
\alpha U_b\!\left(B_{\mathrm{new}}(\Gamma_{\mathrm{in}}^+)\right),
\qquad
B(x)
\ge
L_b\!\left(B(\Gamma_{\mathrm{in}}^-)\right).
\]
Therefore, a sufficient condition for this subcase is
\begin{align}
\alpha U_b\!\left(B_{\mathrm{new}}(\Gamma_{\mathrm{in}}^+)\right)
-
\lambda L_b\!\left(B(\Gamma_{\mathrm{in}}^-)\right)
\le 0 .
\label{eq:proof_in_interface_condition}
\end{align}
Under \cref{eq:proof_in_interface_condition}, we obtain
\[
\begin{aligned}
V(x^+)-\hat{\lambda}(x)V(x)
&=
\alpha B_{\mathrm{new}}(x^+)-\lambda B(x)\\
&\le
\alpha U_b\!\left(B_{\mathrm{new}}(\Gamma_{\mathrm{in}}^+)\right)
-
\lambda L_b\!\left(B(\Gamma_{\mathrm{in}}^-)\right)\\
&\le 0 .
\end{aligned}
\]

\textbf{Case 2: \(x\in S_{\mathrm{new}}\).}
In this case,
\[
V(x)=\alpha B_{\mathrm{new}}(x),\qquad
\hat{\pi}(x)=\pi_{\mathrm{new}}(x),\qquad
\hat{\lambda}(x)=\lambda',
\]
and hence
\[
x^+=f(x,\pi_{\mathrm{new}}(x)).
\]
Again, we distinguish two subcases according to the region containing \(x^+\).

\emph{Subcase 2.1: \(x^+\in S_{\mathrm{new}}\).}
Then \(V(x^+)=\alpha B_{\mathrm{new}}(x^+)\). Since \(B_{\mathrm{new}}\) satisfies the
\cref{eq:bc2-inv} on \(S_{\mathrm{new}}\setminus S_u^{\mathrm{new}}\), we have
\[
B_{\mathrm{new}}(f(x,\pi_{\mathrm{new}}(x)))
-
\lambda' B_{\mathrm{new}}(x)
\le 0 .
\]
Therefore,
\[
V(x^+)-\hat{\lambda}(x)V(x)
=
\alpha B_{\mathrm{new}}(x^+)- \alpha \lambda'B_{\mathrm{new}}(x)
\le 0 .
\]

\emph{Subcase 2.2: \(x^+\in S_{\mathrm{origin}}\).}
Then the transition crosses from \(S_{\mathrm{new}}\) to \(S_{\mathrm{origin}}\). By the definition of the switching interface sets,
\[
x\in \Gamma_{\mathrm{out}}^-,
\qquad
x^+\in \Gamma_{\mathrm{out}}^+ .
\]
In this case,
\[
V(x)=\alpha B_{\mathrm{new}}(x),\qquad V(x^+)=B(x^+),
\]
so the \cref{eq:condition3} becomes
\[
B(x^+)-\alpha \lambda' B_{\mathrm{new}}(x)\le 0 .
\]
Using the upper and lower bounds over the interface sets, we have
\[
B(x^+)
\le
U_b\!\left(B(\Gamma_{\mathrm{out}}^+)\right),
\qquad
\alpha B_{\mathrm{new}}(x)
\ge
\alpha L_b\!\left(B_{\mathrm{new}}(\Gamma_{\mathrm{out}}^-)\right).
\]
Therefore, a sufficient condition for this subcase is
\begin{align}
U_b\!\left(B(\Gamma_{\mathrm{out}}^+)\right)
-
\alpha \lambda' L_b\!\left(B_{\mathrm{new}}(\Gamma_{\mathrm{out}}^-)\right)
\le 0 .
\label{eq:proof_out_interface_condition}
\end{align}
Under \cref{eq:proof_out_interface_condition}, we obtain
\[
\begin{aligned}
V(x^+)-\hat{\lambda}(x)V(x)
&=
B(x^+)-\alpha \lambda'B_{\mathrm{new}}(x)\\
&\le
U_b\!\left(B(\Gamma_{\mathrm{out}}^+)\right)
-
\alpha \lambda' L_b\!\left(B_{\mathrm{new}}(\Gamma_{\mathrm{out}}^-)\right)\\
&\le 0 .
\end{aligned}
\]

Since \cref{eq:proof_in_interface_condition} and \cref{eq:proof_out_interface_condition} are the given condition in \cref{theorem:apwbc}, we have
\[
V(f(x,\hat{\pi}(x)))-\hat{\lambda}(x)V(x)\le 0,
\qquad
\forall x\in S\setminus \hat{S}_u .
\]
Therefore, \cref{eq:condition3} in \cref{df:pwBC} holds.

So, \(V\) satisfies \cref{eq:condition1}, \cref{eq:condition2}, and \cref{eq:condition3} in \cref{df:pwBC}. It means that the adaptive piecewise barrier certificate is a piecewise barrier certificate in \cref{df:pwBC}. By \cref{thm:piecewise}, the adaptive barrier certificate \(V\) can ensure the safety of $\hat{M}$.
\end{proof}

\subsection{Compositional Safety under Sequentially Introduced Obstacles}
\label{app:compositional_safety}

We now show that the adaptive piecewise barrier certificate construction can be
applied sequentially when new unsafe regions are introduced one after another.
Suppose that the first newly introduced unsafe set \(S_u^{\mathrm{new}}\) has
already been repaired. After this repair, the state space is decomposed as
\[
S = S_{\mathrm{origin}} \cup S_{\mathrm{new}},
\qquad
S_{\mathrm{origin}}\cap S_{\mathrm{new}}=\emptyset,
\]
and the certified piecewise barrier certificate is
\[
V(x)=
\begin{cases}
B(x), & x\in S_{\mathrm{origin}},\\
\alpha B_{\mathrm{new}}(x), & x\in S_{\mathrm{new}}.
\end{cases}
\]
Here \(B\) is the original barrier certificate retained on the unaffected region,
and \(B_{\mathrm{new}}\) is the repaired local barrier certificate on
\(S_{\mathrm{new}}\).

Now suppose that another unsafe set, denoted by \(S_u^{\mathrm{new}'}\), is
introduced after the first repair. Let \(S_{\mathrm{new}}'\) be its local repair
region, with
\[
S_u^{\mathrm{new}'}\subseteq S_{\mathrm{new}}'
\subseteq S_{\mathrm{origin}}\setminus S_u .
\]
That is, the new repair region is carved out from the current unaffected region.
We define the remaining unaffected region as
\[
S_{\mathrm{origin}}'
:=
S_{\mathrm{origin}}\setminus S_{\mathrm{new}}' .
\]
Then the state space admits the refined decomposition
\[
S
=
S_{\mathrm{origin}}'
\cup S_{\mathrm{new}}
\cup S_{\mathrm{new}}',
\]
where the three regions are pairwise disjoint. The previously repaired local
certificate \(B_{\mathrm{new}}\) on \(S_{\mathrm{new}}\) is retained, and we only
synthesize a new local certificate \(B_{\mathrm{new}}'\) on
\(S_{\mathrm{new}}'\).

We assume that there are no direct one-step transitions between the two local
repair regions. Formally,
\[
\{x\in S_{\mathrm{new}}
\mid
f(x,\pi_{\mathrm{new}}(x))\in S_{\mathrm{new}}'\}
=
\emptyset,
\]
and
\[
\{x\in S_{\mathrm{new}}'
\mid
f(x,\pi_{\mathrm{new}}'(x))\in S_{\mathrm{new}}\}
=
\emptyset.
\]
Under this assumption, the new repair region only needs to be checked against
the refined unaffected region \(S_{\mathrm{origin}}'\), since all cross-region
transitions involving \(S_{\mathrm{new}}\) have already been handled by the
previous repair.

For the newly introduced repair region \(S_{\mathrm{new}}'\), define the
switching interface sets as
\[
\Gamma_{\mathrm{in}}^{\prime -}
:=
\{x \mid x\in \mathcal H (\Omega(S_0))\cap S_{\mathrm{origin}}',
f(x,\pi(x))\in S_{\mathrm{new}}'\},
\]
\[
\Gamma_{\mathrm{in}}^{\prime +}
:=
\{f(x,\pi(x))\mid x\in \Gamma_{\mathrm{in}}^{\prime -}\},
\]
\[
\Gamma_{\mathrm{out}}^{\prime -}
:=
\{x\mid x\in \mathcal H (\Omega(\Gamma_{\mathrm{in}}^{\prime +}))
\cap S_{\mathrm{new}}',
f(x,\pi_{\mathrm{new}}'(x))\in S_{\mathrm{origin}}'\},
\]
\[
\Gamma_{\mathrm{out}}^{\prime +}
:=
\{f(x,\pi_{\mathrm{new}}'(x))
\mid x\in \Gamma_{\mathrm{out}}^{\prime -}\}.
\]
The entry set of the new local repair region is
\[
S_{0,\mathrm{new}}'=\Gamma_{\mathrm{in}}^{\prime +}.
\]

\begin{theorem}[Compositional Safety]
\label{thm:compositional_safety}
Assume that \(V\) is a valid adaptive piecewise barrier certificate for the
unsafe set
\[
\hat{S}_u = S_u\cup S_u^{\mathrm{new}}.
\]
Suppose that another unsafe set \(S_u^{\mathrm{new}'}\) is introduced with a
repair region \(S_{\mathrm{new}}'\subseteq S_{\mathrm{origin}}\setminus S_u\),
and let
\[
\tilde{S}_u
:=
S_u\cup S_u^{\mathrm{new}}\cup S_u^{\mathrm{new}'}.
\]
Assume that there are no direct one-step transitions between
\(S_{\mathrm{new}}\) and \(S_{\mathrm{new}}'\). If there exists a local barrier
certificate \(B_{\mathrm{new}}'\) on \(S_{\mathrm{new}}'\) under policy
\(\pi_{\mathrm{new}}'\) such that
\begin{align}
&B_{\mathrm{new}}'(x)\le 0 &&
\qquad
\forall x\in S_{0,\mathrm{new}}'\\
&B_{\mathrm{new}}'(x)>0 &&
\qquad
\forall x\in S_u^{\mathrm{new}'}\\
&B_{\mathrm{new}}'(f(x,\pi_{\mathrm{new}}'(x)))
-
\lambda'' B_{\mathrm{new}}'(x)
\le 0 &&
\qquad
\forall x\in S_{\mathrm{new}}'\setminus S_u^{\mathrm{new}'} \label{eq:inv_new}
\end{align}

where \(\lambda''\in[0,1]\), and if the new interface sets satisfy
\begin{align}\label{eq:new_interface}
\beta U_b\bigl(B_{\mathrm{new}}'(\Gamma_{\mathrm{in}}^{\prime +})\bigr)
-
\lambda L_b\bigl(B(\Gamma_{\mathrm{in}}^{\prime -})\bigr)
\le 0 ,
U_b\bigl(B(\Gamma_{\mathrm{out}}^{\prime +})\bigr)
-
\beta \lambda'' L_b\bigl(B_{\mathrm{new}}'(\Gamma_{\mathrm{out}}^{\prime -})\bigr)
\le 0
\end{align}

then the refined piecewise certificate
\[
\tilde{V}(x)=
\begin{cases}
B(x), & x\in S_{\mathrm{origin}}',\\
\alpha B_{\mathrm{new}}(x), & x\in S_{\mathrm{new}},\\
\beta B_{\mathrm{new}}'(x), & x\in S_{\mathrm{new}}',
\end{cases}
\]
together with the piecewise policy
\[
\tilde{\pi}(x)=
\begin{cases}
\pi(x), & x\in S_{\mathrm{origin}}',\\
\pi_{\mathrm{new}}(x), & x\in S_{\mathrm{new}},\\
\pi_{\mathrm{new}}'(x), & x\in S_{\mathrm{new}}',
\end{cases}
\]
ensures safety with respect to the updated unsafe set \(\tilde{S}_u\).
\end{theorem}

\begin{proof}
We prove that \(\tilde{V}\) satisfies the barrier certificate conditions over
the refined three-region decomposition.

First, \cref{eq:bc-init} holds. Since
\(S_0\subseteq S_{\mathrm{origin}}'\) and \(B\) is the original barrier
certificate, we have
\[
\tilde{V}(x)=B(x)\le 0,
\qquad
\forall x\in S_0.
\]

Second, \cref{eq:bc-sep} holds for all unsafe regions. For the original
unsafe set \(S_u\), we have
\[
\tilde{V}(x)=B(x)>0,
\qquad
\forall x\in S_u.
\]
For the first repaired unsafe set \(S_u^{\mathrm{new}}\), the validity of the
previous repair gives
\[
\tilde{V}(x)=\alpha B_{\mathrm{new}}(x)>0,
\qquad
\forall x\in S_u^{\mathrm{new}}.
\]
For the newly introduced unsafe set \(S_u^{\mathrm{new}'}\), the new local
certificate gives
\[
\tilde{V}(x)=\beta B_{\mathrm{new}}'(x)>0,
\qquad
\forall x\in S_u^{\mathrm{new}'}.
\]
Therefore,
\[
\tilde{V}(x)>0,
\qquad
\forall x\in \tilde{S}_u.
\]

It remains to prove the \cref{eq:bc-inv}. Let
\[
x\in S\setminus \tilde{S}_u,
\qquad
x^+=f(x,\tilde{\pi}(x)).
\]
We consider the region containing \(x\).

\textbf{Case 1: \(x\in S_{\mathrm{origin}}'\).}
In this case,
\[
\tilde{V}(x)=B(x),
\qquad
\tilde{\pi}(x)=\pi(x).
\]
If \(x^+\in S_{\mathrm{origin}}'\), then it gives
\[
\tilde{V}(x^+)-\lambda \tilde{V}(x)
=
B(x^+)-\lambda B(x)
\le 0.
\]
If \(x^+\in S_{\mathrm{new}}\), then this transition is already covered by the
previous \cref{eq:interface_set} between \(B\) and \(B_{\mathrm{new}}\). Hence,
\[
\tilde{V}(x^+)-\lambda \tilde{V}(x)
=
\alpha B_{\mathrm{new}}(x^+)-\lambda B(x)
\le 0.
\]
If \(x^+\in S_{\mathrm{new}}'\), then
\(x\in \Gamma_{\mathrm{in}}^{\prime -}\) and
\(x^+\in \Gamma_{\mathrm{in}}^{\prime +}\). By \cref{eq:new_interface},
\[
\begin{aligned}
\tilde{V}(x^+)-\lambda \tilde{V}(x)
&=
\beta B_{\mathrm{new}}'(x^+)-\lambda B(x)\\
&\le
\beta U_b\bigl(B_{\mathrm{new}}'(\Gamma_{\mathrm{in}}^{\prime +})\bigr)
-
\lambda L_b\bigl(B(\Gamma_{\mathrm{in}}^{\prime -})\bigr)
\le 0.
\end{aligned}
\]

\textbf{Case 2: \(x\in S_{\mathrm{new}}\).}
In this case,
\[
\tilde{V}(x)=\alpha B_{\mathrm{new}}(x),
\qquad
\tilde{\pi}(x)=\pi_{\mathrm{new}}(x).
\]
If \(x^+\in S_{\mathrm{new}}\), then \cref{eq:bc2-inv} of
\(B_{\mathrm{new}}\) gives
\[
\tilde{V}(x^+)-\lambda' \tilde{V}(x)
=
\alpha B_{\mathrm{new}}(x^+)-\alpha \lambda' B_{\mathrm{new}}(x)
\le 0.
\]
If \(x^+\in S_{\mathrm{origin}}'\), then this transition is already covered by
the previous \cref{eq:interface_set} between \(B_{\mathrm{new}}\) and \(B\).
Hence,
\[
\tilde{V}(x^+)-\lambda' \tilde{V}(x)
=
B(x^+)-\alpha \lambda' B_{\mathrm{new}}(x)
\le 0.
\]
The remaining case \(x^+\in S_{\mathrm{new}}'\) is excluded by the assumption
that there is no direct transition from \(S_{\mathrm{new}}\) to
\(S_{\mathrm{new}}'\).

\textbf{Case 3: \(x\in S_{\mathrm{new}}'\).}
In this case,
\[
\tilde{V}(x)=\beta B_{\mathrm{new}}'(x),
\qquad
\tilde{\pi}(x)=\pi_{\mathrm{new}}'(x).
\]
If \(x^+\in S_{\mathrm{new}}'\), then \cref{eq:inv_new} of
\(B_{\mathrm{new}}'\) gives
\[
\tilde{V}(x^+)-\lambda'' \tilde{V}(x)
=
\beta B_{\mathrm{new}}'(x^+)- \beta \lambda'' B_{\mathrm{new}}'(x)
\le 0.
\]
If \(x^+\in S_{\mathrm{origin}}'\), then
\(x\in \Gamma_{\mathrm{out}}^{\prime -}\) and
\(x^+\in \Gamma_{\mathrm{out}}^{\prime +}\). By \cref{eq:new_interface},
\[
\begin{aligned}
\tilde{V}(x^+)-\lambda'' \tilde{V}(x)
&=
B(x^+)-\beta \lambda''B_{\mathrm{new}}'(x)\\
&\le
U_b\bigl(B(\Gamma_{\mathrm{out}}^{\prime +})\bigr)
-
\beta \lambda'' L_b\bigl(B_{\mathrm{new}}'(\Gamma_{\mathrm{out}}^{\prime -})\bigr)
\le 0.
\end{aligned}
\]
The remaining case \(x^+\in S_{\mathrm{new}}\) is excluded by the assumption
that there is no direct transition from \(S_{\mathrm{new}}'\) to
\(S_{\mathrm{new}}\).

Combining the three cases, we have shown that
\[
\tilde{V}(f(x,\tilde{\pi}(x)))
-
\tilde{\lambda}(x)\tilde{V}(x)
\le 0,
\qquad
\forall x\in S\setminus \tilde{S}_u,
\]
where
\[
\tilde{\lambda}(x)=
\begin{cases}
\lambda, & x\in S_{\mathrm{origin}}',\\
\lambda', & x\in S_{\mathrm{new}},\\
\lambda'', & x\in S_{\mathrm{new}}'.
\end{cases}
\]
Thus, \(\tilde{V}\) satisfies \cref{eq:bc-init} \cref{eq:bc-sep} \cref{eq:bc-inv} with respect to the updated unsafe set
\(\tilde{S}_u\). Therefore, all trajectories starting from \(S_0\) avoid
\(\tilde{S}_u\), and the system remains safe after introducing
\(S_u^{\mathrm{new}'}\).
\end{proof}

\begin{remark}
If multiple newly introduced obstacles are overlapping, adjacent, or dynamically coupled, they can be conservatively treated as a single enlarged unsafe region. More generally, non-box or irregular obstacle shapes can be over-approximated by boxes or unions of boxes, after which our repair procedure applies to the resulting unsafe region.   
\end{remark}

\begin{remark}
 We do not explicitly address perception
uncertainty, model mismatch, or noisy state estimation, which are
orthogonal to the certificate-repair problem studied here.
When explicit uncertainty bounds are available, they can be incorporated by enlarging the unsafe set or tightening the Lipschitz/discretization margins.
\end{remark}

\subsection{Computation of the Interface Bounds}
\label{app:interface bound}
After obtaining a repaired local barrier certificate, we compute the switching interface sets 
\(\Gamma_{\mathrm{in}}^{-}\), \(\Gamma_{\mathrm{in}}^{+}\), 
\(\Gamma_{\mathrm{out}}^{-}\), and \(\Gamma_{\mathrm{out}}^{+}\) 
and then verify \cref{eq:interface_set}. 
This condition requires four bounds:
\[
U_b(B(\Gamma_{\mathrm{out}}^+)),\qquad
L_b(B_{\mathrm{new}}(\Gamma_{\mathrm{out}}^-)),\qquad
L_b(B(\Gamma_{\mathrm{in}}^-)),\qquad
U_b(B_{\mathrm{new}}(\Gamma_{\mathrm{in}}^+)).
\]

We compute these bounds by discretizing each interface set. 
For an interface set \(G\), let \(\mathcal{D}_G\) be its discretized point set under grid granularity \(\eta_G\). 
That is, every state \(x\in G\) has a discretized representative \(\hat{x}\in \mathcal{D}_G\) satisfying
\[
\|x-\hat{x}\|\le \eta_G .
\]
Let \(C\) denote the barrier certificate evaluated on this set, where \(C\) is either \(B\) or \(B_{\mathrm{new}}\). 
If \(C\) is Lipschitz continuous with constant \(L_C\), then for any \(x\in G\) and its representative \(\hat{x}\in\mathcal{D}_G\),
\[
|C(x)-C(\hat{x})|
\le
L_C\|x-\hat{x}\|
\le
L_C\eta_G .
\]
Therefore, for a discretized point \(\hat{x}\), the value \(C(\hat{x})\) represents all continuous states within distance \(\eta_G\) up to an error of at most \(L_C\eta_G\). 
Thus, the upper and lower bounds on \(C(G)\) are estimated conservatively by
\[
U_b(C(G))
=
\max_{\hat{x}\in\mathcal{D}_G} C(\hat{x})
+
L_C\eta_G,
\]
and
\[
L_b(C(G))
=
\min_{\hat{x}\in\mathcal{D}_G} C(\hat{x})
-
L_C\eta_G .
\]

Accordingly, the four bounds used in \cref{eq:interface_set} are computed as
\[
U_b(B(\Gamma_{\mathrm{out}}^+))
=
\max_{\hat{x}\in\mathcal{D}_{\Gamma_{\mathrm{out}}^+}}
B(\hat{x})
+
L_B\eta_{\Gamma_{\mathrm{out}}^+},
\]
\[
L_b(B_{\mathrm{new}}(\Gamma_{\mathrm{out}}^-))
=
\min_{\hat{x}\in\mathcal{D}_{\Gamma_{\mathrm{out}}^-}}
B_{\mathrm{new}}(\hat{x})
-
L_{B_{\mathrm{new}}}\eta_{\Gamma_{\mathrm{out}}^-},
\]
\[
L_b(B(\Gamma_{\mathrm{in}}^-))
=
\min_{\hat{x}\in\mathcal{D}_{\Gamma_{\mathrm{in}}^-}}
B(\hat{x})
-
L_B\eta_{\Gamma_{\mathrm{in}}^-},
\]
and
\[
U_b(B_{\mathrm{new}}(\Gamma_{\mathrm{in}}^+))
=
\max_{\hat{x}\in\mathcal{D}_{\Gamma_{\mathrm{in}}^+}}
B_{\mathrm{new}}(\hat{x})
+
L_{B_{\mathrm{new}}}\eta_{\Gamma_{\mathrm{in}}^+}.
\]

Finally, these four conservative bounds are substituted into \cref{eq:interface_set} to check whether the repaired local certificate is compatible with the original certificate across the switching interface sets.

\subsection{Box-Shaped Construction of the Local Repair Region}
\label{app:box-shaped-repair-region}

In \cref{sec:BC transfer}, the local repair region is introduced in the form of a Minkowski sum:
\[
S_{\mathrm{new}} = S_u^{\mathrm{new}}\oplus \mathcal{N},
\]
where \(\mathcal{N}\) specifies the neighborhood around the newly introduced unsafe set. In our experiments, we instantiate \(\mathcal{N}\) as a box-shaped neighborhood. This construction is an implementation choice for the benchmarks, while the proposed framework itself only requires
\[
S_u^{\mathrm{new}}\subseteq S_{\mathrm{new}}\subseteq S .
\]

For each target obstacle, suppose the newly introduced unsafe set is represented as an axis-aligned box
\[
S_u^{\mathrm{new}}
=
\prod_{i=1}^{n}
[\underline{x}_i^u,\overline{x}_i^u],
\]
where \(\underline{x}_i^u\) and \(\overline{x}_i^u\) denote the lower and upper bounds of the obstacle along the \(i\)-th state dimension. Let
\[
d_i=\overline{x}_i^u-\underline{x}_i^u
\]
be the obstacle size along this dimension. Given two expansion factors \(\rho_1 , \rho_2>0\), we choose the neighborhood set as
\[
\mathcal{N}(\rho_1,\rho_2)
=
\prod_{i=1}^{n}
[-\rho_1 d_i,\rho_2 d_i].
\]
In general, the two expansion factors is not equal, which means flexible construction of $S_{\mathrm{new}}$. In our experiments, we use $\rho_1=\rho_2$ to simplify, which can be written as 
\[
\mathcal{N}(\rho)
=
\prod_{i=1}^{n}
[-\rho d_i,\rho d_i].
\]

Therefore, the local repair region becomes
\[
S_{\mathrm{new}}(\rho)
=
S_u^{\mathrm{new}}\oplus \mathcal{N}(\rho).
\]

Since both \(S_u^{\mathrm{new}}\) and \(\mathcal{N}(\rho)\) are axis-aligned boxes, their Minkowski sum can be written explicitly as
\[
S_{\mathrm{new}}(\rho)
=
\prod_{i=1}^{n}
[\underline{x}_i^u-\rho d_i,\overline{x}_i^u+\rho d_i].
\]
If the expanded box exceeds the original state space \(S\), it is clipped to \(S\) so that \(S_{\mathrm{new}}(\rho)\subseteq S\).

In our main experiments, we use \(\rho=2\) as the default expansion factor. For example, if the newly introduced unsafe set occupies the interval \([2,3]\) along one state dimension, then \(d=1\), and the corresponding local repair interval with \(\rho=2\) is
\[
[2-2d,\,3+2d]=[0,5].
\]

The expansion factor is not unique and is not a theoretical requirement of the framework. In practice, the repair region can be selected progressively: one may start from a compact region with a smaller \(\rho\), attempt local repair and compositional verification, and enlarge \(\rho\) if the repair or verification fails. This small-to-large strategy keeps the repair problem localized whenever possible, while still allowing additional room for repair when the initial local region is too restrictive.

\subsection{Proof of ~\cref{thm:LipBC}}
\label{app:proof_thm3}
In this section, we prove ~\cref{thm:LipBC}, which shows that the discretized transfer constraints imply the continuous transferred-certificate constraints.

The continuous constraints in
\cref{eq:continuous_init,eq:continuous_unsafe,eq:continuous_dyn}
impose infinitely many inequalities on the unknown last-layer parameters.
Spatial discretization yields finitely many constraints for optimization,
but enforcing them only at representative points does not ensure
validity throughout the state space.
To bridge this gap, the margins
$\delta_{\mathrm{init}}$, $\delta_{\mathrm{unsafe}}$, and
$\delta_{\mathrm{dyn}}$ bound the variation of the respective
constraint functions within distance $\eta$ of each representative.
As derived below, the first two depend on
the Lipschitz constants of the alignment and feature maps,
while the third also accounts for the closed-loop dynamics and decay term.
The bound $\|w_j\|_1\le W$ ensures that these variation bounds
hold uniformly over all admissible last-layer parameters.
Thus, satisfying the tightened finite constraints in
\cref{eq:maindisc_init,eq:maindisc_unsafe,eq:maindisc_dyn}
guarantees the original continuous constraints
by \cref{thm:LipBC}.
A smaller $\eta$ reduces the required margins
but increases the number of representative points.

For a fixed policy parameter $\theta$ and prescribed margins,
the discretized constraints are linear in $w_j$ and can be
stacked as a linear system $A(\theta)w_j\le-b$.

To see this explicitly, let
\[
\mathcal{D}_{0}^{j}=\{\hat{x}_{0,r}\}_{r=1}^{N_0},\quad
\mathcal{D}_{u}^{j}=\{\hat{x}_{u,r}\}_{r=1}^{N_u},\quad
\mathcal{D}_{\mathrm{dyn}}^{j}=\{\hat{x}_{d,r}\}_{r=1}^{N_d}.
\]
For these discretized state sets, define
\begin{align}
A_{\mathrm{init}}
&=
\begin{bmatrix}
\phi(T_{ji}(\hat{x}_{0,1}))^\top\\
\vdots\\
\phi(T_{ji}(\hat{x}_{0,N_0}))^\top
\end{bmatrix},
\\
A_{\mathrm{unsafe}}
&=
\begin{bmatrix}
-\phi(T_{ji}(\hat{x}_{u,1}))^\top\\
\vdots\\
-\phi(T_{ji}(\hat{x}_{u,N_u}))^\top
\end{bmatrix},
\\
A_{\mathrm{dyn}}(\theta)
&=
\begin{bmatrix}
\left(\phi(T_{ji}(f(\hat{x}_{d,1},\pi(\hat{x}_{d,1};\theta))))-\lambda'\phi(T_{ji}(\hat{x}_{d,1}))\right)^\top\\
\vdots\\
\left(\phi(T_{ji}(f(\hat{x}_{d,N_d},\pi(\hat{x}_{d,N_d};\theta))))-\lambda'\phi(T_{ji}(\hat{x}_{d,N_d}))\right)^\top
\end{bmatrix}.
\end{align}
Then all discretized constraints can be stacked as
\begin{align}
A(\theta)w_j \le -b,
\label{eq:target}
\end{align}
where
\begin{align}
A(\theta)
=
\begin{bmatrix}
A_{\mathrm{init}}\\
A_{\mathrm{unsafe}}\\
A_{\mathrm{dyn}}(\theta)
\end{bmatrix},
\qquad
b=
\begin{bmatrix}
\delta_{\mathrm{init}}\mathbf{1}_{N_0}\\
(\delta_{\mathrm{unsafe}}+\beta)\mathbf{1}_{N_u}\\
\delta_{\mathrm{dyn}}\mathbf{1}_{N_d}
\end{bmatrix}.
\end{align}
Here, each row of \(A(\theta)\) corresponds to one discretized transfer constraint, and $b$ collects the prescribed margins. The dependence on \(\theta\) appears only in \(A_{\mathrm{dyn}}(\theta)\), because the local decrease constraints involve the successor state \(f(x,\pi(x;\theta))\). Therefore, finding a repaired local barrier certificate is reduced to finding a policy parameter \(\theta\) and a final-layer parameter \(w_j\) satisfying \cref{eq:target}.

\textbf{Theorem~\ref{thm:LipBC}} (Continuous validity from discretized transfer constraints)
If a function $B_j$ satisfies that 
\begin{align}
g_{\mathrm{init}}(\hat{x};w_j)
&\le -\delta_{\mathrm{init}},
&& \forall \hat{x}\in \mathcal{D}_{0}^{j},
\tag{\ref{eq:maindisc_init}}
\\
g_{\mathrm{unsafe}}(\hat{x};w_j)
&\le -(\delta_{\mathrm{unsafe}}+\beta),
&& \forall \hat{x}\in \mathcal{D}_{u}^{j},
\tag{\ref{eq:maindisc_unsafe}}
\\
g_{\mathrm{dyn}}(\hat{x};w_j,\theta)
&\le -\delta_{\mathrm{dyn}},
&& \forall \hat{x}\in \mathcal{D}_{\mathrm{dyn}}^{j},
\tag{\ref{eq:maindisc_dyn}}
\end{align}
where constants
$\delta_{\mathrm{init}}$, \(\delta_{\mathrm{unsafe}}\), and \(\delta_{\mathrm{dyn}}\) 
are derived from the granularity \(\eta\) and the relevant Lipschitz constants of the system, then $B_j$ is a valid transferred BC on $S^j$ that satisfies the continuous constraints in~\cref{eq:continuous_init,eq:continuous_unsafe,eq:continuous_dyn}.

\begin{proof}
Let \(\eta>0\) be the grid granularity for discretization. For each continuous state \(x\) in the corresponding domain, there exists a discretized representative \(\hat{x}\) such that
\[
\|x-\hat{x}\|_1 \le \eta .
\]
This produces finite discretized state sets 
\(\mathcal{D}_{0}^{j}\subseteq S_{0}^j\), \(\mathcal{D}_{u}^{j}\subseteq S_{u}^{j}\), and \(\mathcal{D}_{\mathrm{dyn}}^{j}\subseteq S^j\) 
for the entry set, the newly introduced unsafe set, and the local repair region, respectively. 
Assume that the final-layer parameter is bounded by some constant $W>0$, i.e., 
\[
\|w_j\|_1\le W,
\]
and that the alignment map \(T_{ji}\), the feature map \(\phi\), the system dynamics \(f\), and the policy class \(\pi(\cdot;\theta)\) are Lipschitz continuous with constants \(L_T\), \(L_\phi\), \(L_f\), and \(\bar{L}_\pi\), respectively. The discretization margins are chosen as
\[
\delta_{\mathrm{init}} = W L_\phi L_T \eta,
\]
\[
\delta_{\mathrm{unsafe}} = W L_\phi L_T \eta,
\]
and
\[
\delta_{\mathrm{dyn}}
=
W L_\phi L_T
\left(
L_f(1+\bar{L}_\pi)+\lambda'
\right)\eta .
\]
We prove the three continuous constraints separately.

First, consider an arbitrary state \(x\in S_{0}^j\). By the discretization construction, there exists \(\hat{x}\in \mathcal{D}_{0}^{j}\) such that
\[
\|x-\hat{x}\|_1 \le \eta .
\]
By the definition of \(g_{\mathrm{init}}\),
\[
g_{\mathrm{init}}(x;w_j)-g_{\mathrm{init}}(\hat{x};w_j)
=
w_j^\top
\left(
\phi(T_{ji}(x))-\phi(T_{ji}(\hat{x}))
\right).
\]
Using Hölder's inequality, we have
\[
\left|
g_{\mathrm{init}}(x;w_j)-g_{\mathrm{init}}(\hat{x};w_j)
\right|
\le
\|w_j\|_1
\left\|
\phi(T_{ji}(x))-\phi(T_{ji}(\hat{x}))
\right\|_\infty .
\]
By the Lipschitz continuity of \(\phi\) and \(T_{ji}\),
\[
\left\|
\phi(T_{ji}(x))-\phi(T_{ji}(\hat{x}))
\right\|_\infty
\le
L_\phi L_T \|x-\hat{x}\|_1 .
\]
Therefore,
\[
\left|
g_{\mathrm{init}}(x;w_j)-g_{\mathrm{init}}(\hat{x};w_j)
\right|
\le
W L_\phi L_T \|x-\hat{x}\|
\le
W L_\phi L_T \eta
=
\delta_{\mathrm{init}} .
\]
Since the discretized entry constraint in \cref{eq:maindisc_init} gives
\[
g_{\mathrm{init}}(\hat{x};w_j)\le -\delta_{\mathrm{init}},
\]
we obtain
\[
g_{\mathrm{init}}(x;w_j)
\le
g_{\mathrm{init}}(\hat{x};w_j)+\delta_{\mathrm{init}}
\le
-\delta_{\mathrm{init}}+\delta_{\mathrm{init}}
=
0 .
\]
Because \(x\in S_{0}^j\) was chosen arbitrarily,
\[
g_{\mathrm{init}}(x;w_j)\le 0,
\qquad
\forall x\in S_{0,\mathrm{new}}^j .
\]
Thus, the continuous entry constraint in \cref{eq:continuous_init} holds.

Second, let \(x\in S_u^j\) be chosen arbitrarily. By the discretization construction, there exists \(\hat{x}\in \mathcal{D}_{u}^{j}\) such that
\[
\|x-\hat{x}\|_1 \le \eta .
\]
By the definition of \(g_{\mathrm{unsafe}}\),
\[
g_{\mathrm{unsafe}}(x;w_j)-g_{\mathrm{unsafe}}(\hat{x};w_j)
=
-w_j^\top
\left(
\phi(T_{ji}(x))-\phi(T_{ji}(\hat{x}))
\right).
\]
Repeating the same Lipschitz argument as above yields
\[
\left|
g_{\mathrm{unsafe}}(x;w_j)-g_{\mathrm{unsafe}}(\hat{x};w_j)
\right|
\le
W L_\phi L_T \|x-\hat{x}\|_1
\le
W L_\phi L_T \eta
=
\delta_{\mathrm{unsafe}} .
\]
Since the discretized unsafe-set constraint in \cref{eq:maindisc_unsafe} gives
\[
g_{\mathrm{unsafe}}(\hat{x};w_j)
\le
-(\delta_{\mathrm{unsafe}}+\beta),
\]
we obtain
\[
g_{\mathrm{unsafe}}(x;w_j)
\le
g_{\mathrm{unsafe}}(\hat{x};w_j)+\delta_{\mathrm{unsafe}}
\le
-(\delta_{\mathrm{unsafe}}+\beta)+\delta_{\mathrm{unsafe}}
=
-\beta .
\]
Because \(x\in S_u^j\) was chosen arbitrarily,
\[
g_{\mathrm{unsafe}}(x;w_j)\le -\beta,
\qquad
\forall x\in S_u^j .
\]
Thus, the continuous unsafe-set constraint in \cref{eq:continuous_unsafe} holds.

Third, let \(x\in S^j\) was chosen arbitrarily. By the discretization construction, there exists \(\hat{x}\in \mathcal{D}_{\mathrm{dyn}}^{j}\) such that
\[
\|x-\hat{x}\|_1 \le \eta .
\]
Define
\[
z(x,\theta)
:=
\phi(T_{ji}(f(x,\pi(x;\theta))))
-
\lambda'\phi(T_{ji}(x)).
\]
Then
\[
g_{\mathrm{dyn}}(x;w_j,\theta)
=
w_j^\top z(x,\theta).
\]
Hence, by Hölder's inequality,
\[
\left|
g_{\mathrm{dyn}}(x;w_j,\theta)
-
g_{\mathrm{dyn}}(\hat{x};w_j,\theta)
\right|
\le
\|w_j\|_1
\|z(x,\theta)-z(\hat{x},\theta)\|_\infty .
\]
We now bound \(\|z(x,\theta)-z(\hat{x},\theta)\|_\infty\). By the triangle inequality,
\[
\begin{aligned}
\|z(x,\theta)-z(\hat{x},\theta)\|_\infty
\le\;&
\left\|
\phi(T_{ji}(f(x,\pi(x;\theta))))
-
\phi(T_{ji}(f(\hat{x},\pi(\hat{x};\theta))))
\right\|_\infty \\
&+
\lambda'
\left\|
\phi(T_{ji}(x))-\phi(T_{ji}(\hat{x}))
\right\|_\infty .
\end{aligned}
\]
Using the Lipschitz continuity of \(\phi\) and \(T_{ji}\),
\[
\left\|
\phi(T_{ji}(f(x,\pi(x;\theta))))
-
\phi(T_{ji}(f(\hat{x},\pi(\hat{x};\theta))))
\right\|_\infty
\le
L_\phi L_T
\left\|
f(x,\pi(x;\theta))
-
f(\hat{x},\pi(\hat{x};\theta))
\right\|_1 .
\]
Since \(f\) is \(L_f\)-Lipschitz,
\[
\left\|
f(x,\pi(x;\theta))
-
f(\hat{x},\pi(\hat{x};\theta))
\right\|_1
\le
L_f
\left(
\|x-\hat{x}\|_1
+
\|\pi(x;\theta)-\pi(\hat{x};\theta)\|_1
\right).
\]
Using the Lipschitz continuity of the policy class,
\[
\|\pi(x;\theta)-\pi(\hat{x};\theta)\|_1
\le
\bar{L}_\pi \|x-\hat{x}\|_1.
\]
Therefore,
\[
\left\|
f(x,\pi(x;\theta))
-
f(\hat{x},\pi(\hat{x};\theta))
\right\|_1
\le
L_f(1+\bar{L}_\pi)\|x-\hat{x}\|_1.
\]
Moreover,
\[
\left\|
\phi(T_{ji}(x))-\phi(T_{ji}(\hat{x}))
\right\|_\infty
\le
L_\phi L_T \|x-\hat{x}\|_1.
\]
Combining the above inequalities gives
\[
\|z(x,\theta)-z(\hat{x},\theta)\|_\infty
\le
L_\phi L_T
\left(
L_f(1+\bar{L}_\pi)+\lambda'
\right)
\|x-\hat{x}\|_1.
\]
Using \(\|w_j\|_1\le W\), we obtain
\[
\left|
g_{\mathrm{dyn}}(x;w_j,\theta)
-
g_{\mathrm{dyn}}(\hat{x};w_j,\theta)
\right|
\le
W L_\phi L_T
\left(
L_f(1+\bar{L}_\pi)+\lambda'
\right)
\|x-\hat{x}\|_1
\le
W L_\phi L_T
\left(
L_f(1+\bar{L}_\pi)+\lambda'
\right)
\eta
=
\delta_{\mathrm{dyn}} .
\]
Since the discretized decrease constraint in \cref{eq:maindisc_dyn} gives
\[
g_{\mathrm{dyn}}(\hat{x};w_j,\theta)\le -\delta_{\mathrm{dyn}},
\]
we obtain
\[
g_{\mathrm{dyn}}(x;w_j,\theta)
\le
g_{\mathrm{dyn}}(\hat{x};w_j,\theta)+\delta_{\mathrm{dyn}}
\le
-\delta_{\mathrm{dyn}}+\delta_{\mathrm{dyn}}
=
0 .
\]
Because \(x\in S_{\mathrm{new}}^j\) was chosen arbitrarily,
\[
g_{\mathrm{dyn}}(x;w_j,\theta)\le 0,
\qquad
\forall x\in S_{\mathrm{new}}^j .
\]
Thus, the continuous decrease constraint in \cref{eq:continuous_dyn} holds.

Combining the three parts, we conclude that the discretized constraints in \cref{eq:maindisc_init,eq:maindisc_unsafe,eq:maindisc_dyn} imply the continuous constraints in \cref{eq:continuous_init,eq:continuous_unsafe,eq:continuous_dyn}. This proves \cref{thm:LipBC}.
\end{proof}

\subsection{Supplementary Proof for \cref{sec:BC transfer}}
\label{app:sec5}
\begin{theorem}[Continuous formulation of transferred local repair]
Consider the transferred local barrier certificate \(B_j\) defined as $B_j(x)=w_j^\top \phi(T_{ji}(x))$ . If there exists a final-layer parameter \(w_j\) such that the continuous constraints in \cref{eq:continuous_init,eq:continuous_unsafe,eq:continuous_dyn} hold, then \(B_j\) satisfies the local barrier conditions in \cref{eq:bc2-init,eq:bc2-sep,eq:bc2-inv} on \(S_{\mathrm{new}}^j\).
\label{thm:continuous_transfer_repair}
\end{theorem}

\begin{proof}
Consider the \(j\)-th target repair region \(S_{\mathrm{new}}^j\), and let the repaired local barrier certificate be instantiated by the transferred certificate
\[
B_{\mathrm{new}}(x)=B_j(x)=w_j^\top \phi(T_{ji}(x)),
\qquad x\in S_{\mathrm{new}}^j .
\]
We prove that \(B_j\) satisfies the three local barrier conditions.

First, by \cref{eq:continuous_init}, for all \(x\in S_{0,\mathrm{new}}^j\),
\[
g_{\mathrm{init}}(x;w_j)
=
w_j^\top \phi(T_{ji}(x))
=
B_j(x)
\le 0 .
\]
Therefore,
\[
B_{\mathrm{new}}(x)\le 0,
\qquad \forall x\in S_{0,\mathrm{new}}^j,
\]
which is \cref{eq:bc2-init}.

Second, by \cref{eq:continuous_unsafe}, for all \(x\in S_u^j\),
\[
g_{\mathrm{unsafe}}(x;w_j)
=
-w_j^\top \phi(T_{ji}(x))
=
-B_j(x)
\le -\beta .
\]
Since \(\beta>0\), this implies
\[
B_j(x)\ge \beta>0,
\qquad \forall x\in S_u^j .
\]
Hence,
\[
B_{\mathrm{new}}(x)>0,
\qquad \forall x\in S_u^j,
\]
which is \cref{eq:bc2-sep}.

Third, by \cref{eq:continuous_dyn}, for all \(x\in S_{\mathrm{new}}^j\),
\[
g_{\mathrm{dyn}}(x;w_j,\theta)
=
w_j^\top
\left(
\phi(T_{ji}(f(x,\pi(x;\theta))))
-
\lambda'\phi(T_{ji}(x))
\right)
\le 0 .
\]
Using the definition of \(B_j\), this inequality is equivalent to
\[
B_j(f(x,\pi(x;\theta)))-\lambda' B_j(x)\le 0,
\qquad \forall x\in S_{\mathrm{new}}^j .
\]
Let the local policy be \(\pi_{\mathrm{new}}=\pi_{\mathrm{new}}^j(\cdot;\theta)\) on \(S_{\mathrm{new}}^j\). Then
\[
B_{\mathrm{new}}(f(x,\pi_{\mathrm{new}}(x)))
-
\lambda' B_{\mathrm{new}}(x)
\le 0,
\qquad \forall x\in S_{\mathrm{new}}^j .
\]
Since \(S_{\mathrm{new}}^j\setminus S_u^j\subseteq S_{\mathrm{new}}^j\), the same inequality also holds for all \(x\in S_{\mathrm{new}}^j\setminus S_u^j\). This proves the local decrease condition in \cref{eq:bc2-inv}.

Combining these conditions, \(B_j\) satisfies the local barrier conditions in \cref{eq:bc2-init,eq:bc2-sep,eq:bc2-inv} on \(S_{\mathrm{new}}^j\). Therefore, \(B_j\) is a valid repaired local barrier certificate on the target repair region.
\end{proof}

\subsection{Proof of \cref{thm:mainstage1_success}}
\label{app:prove_thm4}

In this section, we prove \cref{thm:mainstage1_success}, which gives the success condition for Stage~I local repair under the original policy.

\textbf{Theorem~\ref{thm:mainstage1_success}} (Feasibility for Stage I).
Assume that \(\mathbf{(P_{\mathrm{I}})}\) attains an optimal solution. Then $\varepsilon_{\mathrm{I}}^\star=0$ if and only if there exists a last-layer parameter \(w_j\) satisfying $A(\theta)w_j\le -b$. Consequently, Stage~I succeeds and yields a valid localized barrier certificate as defined in~\cref{df:apwbc}.

\begin{proof}
Recall that Stage~I fixes the policy parameter as \(\theta=\theta_0\) and solves the following linear program:
\begin{equation}
\begin{aligned}
(\mathbf{P}_{\mathrm{I}})\qquad
\min_{w_j,\xi,u_j}\quad
    & \mathbf{1}^{\top}\xi \\
\text{s.t.}\quad
    & A(\theta_0)w_j\le -b+\xi, \\
    & \xi\ge0, \\
    & -u_j\le w_j\le u_j, \\
    & \mathbf{1}^{\top}u_j\le W, \\
    & u_j\ge0.
\end{aligned}
\label{eq:stageI_full_lp}
\end{equation}
Let $(w_{j,\mathrm{I}}^\star,\xi_{\mathrm{I}}^\star,
u_{j,\mathrm{I}}^\star)$ be an optimal solution
of $(\mathbf{P}_{\mathrm{I}})$,
and let
\[
\varepsilon_{\mathrm{I}}^\star
=
\mathbf{1}^{\top}\xi_{\mathrm{I}}^\star
\]
be the optimal aggregate slack.

We prove the equivalence in two directions.
First, suppose there exists a final-layer parameter
$\bar{w}_j$ such that
\[
A(\theta_0)\bar{w}_j\le-b,
\qquad
\|\bar{w}_j\|_1\le W.
\]

Then choosing
\[
w_j=\bar{w}_j,\qquad \xi=0, \qquad u_j=|\bar{w_j}|
\]
satisfies
\[
A(\theta_0)w_j
=
A(\theta_0)\bar{w}_j
\le -b
=
-b+\xi,
\]
and also satisfies \(\xi\ge 0\). 
Hence, $(\bar{w}_j,0,|\bar{w}_j|)$ is feasible for $(\mathbf{P}_{\mathrm{I}})$, with objective value
$\mathbf{1}^{\top}\xi=0$.

Since every feasible solution of \(\mathbf{(P_{\mathrm{I}})}\) satisfies \(\xi\ge 0\), the objective value \(\mathbf{1}^{\top}\xi\) is always nonnegative. Therefore, the optimal value must be
\[
\varepsilon_{\mathrm{I}}^\star=0 .
\]

Conversely, suppose that
\[
\varepsilon_{\mathrm{I}}^\star=0 .
\]
Because \(\mathbf{(P_{\mathrm{I}})}\) attains an optimal solution, 
there exists $(w_{j,\mathrm{I}}^\star,
\xi_{\mathrm{I}}^\star,u_{j,\mathrm{I}}^\star)$
such that
\[
A(\theta_0)w_{j,\mathrm{I}}^\star
\le
-b+\xi_{\mathrm{I}}^\star,
\qquad
\xi_{\mathrm{I}}^\star\ge 0,
\]
and
\[
\mathbf{1}^{\top}\xi_{\mathrm{I}}^\star=0 .
\]
Since every component of \(\xi_{\mathrm{I}}^\star\) is nonnegative, the equality
\[
\mathbf{1}^{\top}\xi_{\mathrm{I}}^\star=0
\]
implies
\[
\xi_{\mathrm{I}}^\star=0 .
\]
Substituting \(\xi_{\mathrm{I}}^\star=0\) into the relaxed constraint yields
\[
A(\theta_0)w_{j,\mathrm{I}}^\star\le -b .
\]

Moreover, the optimal solution satisfies
$-u_{j,\mathrm{I}}^\star\le
w_{j,\mathrm{I}}^\star\le
u_{j,\mathrm{I}}^\star$ and
$\mathbf{1}^{\top}u_{j,\mathrm{I}}^\star\le W$.
Therefore,
\[
\|w_{j,\mathrm{I}}^\star\|_1
\le
\mathbf{1}^{\top}u_{j,\mathrm{I}}^\star
\le W.
\]

Therefore, there exists a final-layer parameter \(w_j\), namely \(w_{j,\mathrm{I}}^\star\), such that
\[
A(\theta_0)w_j\le -b .
\]

Combining the two directions, we have shown that
\[
    \varepsilon_{\mathrm{I}}^\star=0
    \quad\Longleftrightarrow\quad
    \exists w_j:
    \quad A(\theta_0)w_j\le-b,
    \quad \|w_j\|_1\le W.
\]

Thus, Stage~I succeeds if and only if the optimal aggregate slack is zero.

By \cref{thm:LipBC}, we further have that $B_j(x)=w_j^\top \phi(T_{ji}(x))$ for $x\in S^j$ is a valid transferred BC on $S^j$ that satisfies the continuous constraints in~\cref{eq:continuous_init,eq:continuous_unsafe,eq:continuous_dyn}.   
Then by \cref{thm:continuous_transfer_repair}, we have that \(B_j\) satisfies the local barrier conditions in \cref{eq:bc2-init,eq:bc2-sep,eq:bc2-inv} on \(S_{}^j\) with $S^j=S_{\mathrm{new}}$. Therefore, we conclude that Stage~I succeeds and yields a valid localized barrier certificate as defined in~\cref{df:apwbc}.

\end{proof}

\subsection{Proof of \cref{thm:mainstage2_success}}
\label{app:prove_thm5}
In this subsection, we prove \cref{thm:mainstage2_success}, which gives the success condition for Stage~II local repair when the policy parameter is allowed to change.

\textbf{Theorem~\ref{thm:mainstage2_success}} (Feasibility for Stage II) Assume that \(\mathbf{(P_{\mathrm{II}})}\) attains an optimal solution. Then $\varepsilon_{\mathrm{II}}(\theta)=0$
if and only if there exists a tuple \((w_j,\theta)\) such that $A(\theta)w_j\le -b $.
Consequently, Stage~II succeeds and yields a valid localized barrier certificate as defined in~\cref{df:apwbc}.

\begin{proof}
Fix a policy parameter \(\theta\). Recall that, for this fixed \(\theta\), Stage~II solves the lower-level linear program
\begin{equation}
\begin{aligned}
(\mathbf{P}_{\mathrm{II}}^{\mathrm{low}}(\theta))\qquad
\min_{w_j,\xi,u_j}\quad
    & \mathbf{1}^{\top}\xi \\
\text{s.t.}\quad
    & A(\theta)w_j\le -b+\xi, \\
    & \xi\ge0, \\
    & -u_j\le w_j\le u_j, \\
    & \mathbf{1}^{\top}u_j\le W, \\
    & u_j\ge0.
\end{aligned}
\label{eq:stageII_full_lower_lp}
\end{equation}

Let
\[
(w_{j,\mathrm{II}}^\star(\theta),
\xi_{\mathrm{II}}^\star(\theta),
u_{j,\mathrm{II}}^\star(\theta))
\]
be an optimal solution of \(\mathbf{(P_{\mathrm{II}}^{\mathrm{low}}(\theta))}\). By definition, the optimal aggregate slack is
\[
\varepsilon_{\mathrm{II}}(\theta)
=
\mathbf{1}^{\top}\xi_{\mathrm{II}}^\star(\theta).
\]

We prove the equivalence in two directions. 

First, suppose that there exists a final-layer parameter \(\bar{w}_j\) such that
\[
A(\theta)\bar{w}_j\le -b ,
\qquad
\|\bar{w}_j\|_1\le W.
\]

Then choosing
\[
w_j=\bar{w}_j,\qquad \xi=0, \qquad u_j=|\bar{w_j}|
\]
satisfies the relaxed lower-level constraint:
\[
A(\theta)w_j
=
A(\theta)\bar{w}_j
\le -b
=
-b+\xi,
\]
and also satisfies \(\xi\ge 0\). Hence, $(\bar{w}_j,0,|\bar{w}_j|)$ is feasible
for $(P_{\mathrm{II}}^{\mathrm{low}}(\theta))$,
with objective value
$\mathbf{1}^{\top}\xi=0$.

Since every feasible solution satisfies \(\xi\ge 0\), the objective value \(\mathbf{1}^{\top}\xi\) is always nonnegative. Therefore, the optimal value must be
\[
\varepsilon_{\mathrm{II}}(\theta)=0 .
\]

Conversely, suppose that
\[
\varepsilon_{\mathrm{II}}(\theta)=0 .
\]
Since \(\mathbf{(P_{\mathrm{II}}^{\mathrm{low}}(\theta))}\) attains an optimal solution, there exists
\[
(w_{j,\mathrm{II}}^\star(\theta),\xi_{\mathrm{II}}^\star(\theta),u_{j,\mathrm{II}}^\star(\theta))
\]
such that
\[
A(\theta)w_{j,\mathrm{II}}^\star(\theta)
\le
-b+\xi_{\mathrm{II}}^\star(\theta),
\qquad
\xi_{\mathrm{II}}^\star(\theta)\ge 0,
\]
and
\[
\mathbf{1}^{\top}\xi_{\mathrm{II}}^\star(\theta)=0 .
\]
Because every component of \(\xi_{\mathrm{II}}^\star(\theta)\) is nonnegative, the equality
\[
\mathbf{1}^{\top}\xi_{\mathrm{II}}^\star(\theta)=0
\]
implies
\[
\xi_{\mathrm{II}}^\star(\theta)=0 .
\]
Substituting \(\xi_{\mathrm{II}}^\star(\theta)=0\) into the relaxed constraint gives
\[
A(\theta)w_{j,\mathrm{II}}^\star(\theta)\le -b .
\]

Furthermore, the auxiliary constraints of the
lower-level problem ensure that
\[
\|w_{j,\mathrm{II}}^\star(\theta)\|_1
\le
\mathbf{1}^{\top}
u_{j,\mathrm{II}}^\star(\theta)
\le W.
\]

Therefore, there exists a final-layer parameter \(w_j\), namely \(w_{j,\mathrm{II}}^\star(\theta)\), such that
\[
A(\theta)w_j\le -b .
\]

Combining the two directions, we obtain
\[
    \varepsilon_{\mathrm{II}}(\theta)=0
    \quad\Longleftrightarrow\quad
    \exists w_j:
    \quad A(\theta)w_j\le-b,
    \quad \|w_j\|_1\le W.
\]

Finally, if Stage~II finds a policy parameter \(\theta^\star\) such that
\[
\varepsilon_{\mathrm{II}}(\theta^\star)=0,
\]
then by the equivalence proved above, there exists a corresponding final-layer parameter
\[
w_j^\star = w_{j,\mathrm{II}}^\star(\theta^\star)
\]
such that
\[
A(\theta^\star)w_j^\star\le -b .
\]
This is exactly the compact transfer constraint in \cref{eq:target} under the adapted local policy \(\pi(\cdot;\theta^\star)\).

By \cref{thm:LipBC}, we further have that $B_j(x)=w_j^\top \phi(T_{ji}(x))$ for $x\in S^j$ is a valid transferred BC on $S^j$ that satisfies the continuous constraints in~\cref{eq:continuous_init,eq:continuous_unsafe,eq:continuous_dyn}.   
Then by \cref{thm:continuous_transfer_repair}, we have that \(B_j\) satisfies the local barrier conditions in \cref{eq:bc2-init,eq:bc2-sep,eq:bc2-inv} on \(S_{}^j\) with $S^j=S_{\mathrm{new}}$. Therefore, we conclude that Stage~II succeeds and yields a valid localized barrier certificate as defined in~\cref{df:apwbc}.
\end{proof}

\subsection{Detailed Experimental Results}
\label{app:detailed_results}

This section provides the complete experimental tables corresponding to the compact tables reported in \cref{sec:evaluation}. 
Specifically, ~\cref{tab:full-multi-obstacle} is the complete version of \cref{tab:main_multi_obstacle}, ~\cref{tab:full-region-size} is the complete version of \cref{tab:size_sensitivity}, ~\cref{tab:full-template-selection} is the complete version of \cref{tab:template-selection} and \cref{tab:full_shield_multi_obstacle} is the complete version of \cref{tab:shield_multi_obstacle}. We additionally examine the learning-rate sensitivity of global retraining in \cref{tab:retrain_learning_rate} and discuss the scope of the obstacle-location evaluation.

\paragraph{End-to-End Repair under Multiple Obstacles.}\cref{tab:main_multi_obstacle} in the main text reports a representative subset of the end-to-end running-time results under multiple newly introduced obstacles, while ~\cref{tab:full-multi-obstacle} reports the complete results over all five sequential obstacles for each benchmark. The full table includes the running time of global retraining, the three stages of our repair pipeline, verification time, and the final end-to-end repair time. These complete results further support the same conclusion as \cref{tab:main_multi_obstacle}: global retraining often becomes infeasible under sequentially introduced obstacles, whereas the proposed localized repair pipeline restores safety for all obstacles within the deployment-time budget.

\paragraph{Effect of Repair Region Size.}\cref{tab:size_sensitivity} in the main text reports representative results on the effect of the repair-region size, while ~\cref{tab:full-region-size} provides the complete results across all four benchmarks. These results show the same trade-off consistently across benchmarks: overly small repair regions may make repair infeasible, moderate repair regions often allow efficient Stage~I repair, and overly large repair regions increase the computational cost and may require Stage~II or lead to timeout.

\paragraph{Impact of Template Selection.}\cref{tab:template-selection} in the main text summarizes the impact of template selection on MC and PD, where the effect is most pronounced. ~\cref{tab:full-template-selection} reports the complete template-selection comparison across all four benchmarks. The full results confirm that using a repaired local certificate as the transfer template is generally more effective than repeatedly using the original global certificate, especially in sequential repair settings where later obstacles can benefit from templates adapted to previously repaired regions.

\paragraph{Comparison with Shielding.}\cref{tab:shield_multi_obstacle} in the main text reports a representative comparison with shielding on B1 and B2, while \cref{tab:full_shield_multi_obstacle} provides the complete results across all four benchmarks. The full results support the same conclusion: under repeated executions following persistent environmental changes, our method achieves a
100\% empirical success rate across all evaluated configurations, with consistently lower total runtime and higher average cumulative reward than shielding. The two approaches offer different cost trade-offs:
shielding relies on continued runtime monitoring and intervention
when necessary, whereas our method incurs a one-time repair and
verification cost after each environmental change.
For persistent localized changes, this cost can be amortized over
repeated executions while the updated environment remains unchanged.
Our results therefore demonstrate an advantage in this setting,
rather than uniform superiority over shielding.

\paragraph{Learning-Rate Sensitivity of Global Retraining.}To assess the sensitivity of global retraining to the learning
rate, we evaluate three settings, $10^{-5}$, $10^{-4}$, and
$10^{-3}$, across all configurations,
comprising four benchmarks and five sequential obstacle
configurations per benchmark.
Each configuration is evaluated over
20 runs; $10^{-5}$ is the setting used in the main experiments.
We measure the end-to-end time to obtain a controller--certificate
pair that passes formal verification, including both training
and verification rather than gradient updates alone.
The results are reported in~\cref{tab:retrain_learning_rate}.

At $10^{-5}$, the runtimes for obtaining verified solutions
are of the same order of magnitude as the global-retraining
results reported in Table~1.
In contrast, at $10^{-4}$ and $10^{-3}$, all runs across
the 20 configurations time out without obtaining a verified
safe solution.
Thus, increasing the learning rate does not improve the
observed end-to-end certification outcomes under the evaluated
settings. These results highlight that aggressive parameter updates
do not necessarily translate into a shorter time to obtain
a verified safe controller--certificate pair.

\paragraph{Discussion of Obstacle Locations.}The sequential-obstacle experiments in \cref{tab:full-multi-obstacle} include multiple randomly placed obstacles across the benchmarks,
providing an initial evaluation of certificate repair under
different obstacle locations. However, these experiments do
not isolate the effect of obstacle location on repair difficulty.
A more systematic sensitivity analysis of how obstacle location
affects repair feasibility and runtime is left for future work.

\begin{table*}[!htbp]
\centering
\small
\caption{Running time (in seconds) comparison under multiple newly introduced obstacles across benchmarks.}
\label{tab:full-multi-obstacle}
\resizebox{\textwidth}{!}{
\begin{tabular}{c c c c c c c c c}
\toprule
\multirow{2}{*}{Benchmark}
& \multirow{2}{*}{Original BC}
& \multicolumn{2}{c}{Baseline}
& \multicolumn{5}{c}{Ours} \\
\cmidrule(lr){3-4} \cmidrule(lr){5-9}
& 
& Obstacle ID
& Global retrain
& Stage I
& Stage II
& Stage III
& Verification
& Total time \\
\midrule
\multirow{5}{*}{B1}
& \multirow{5}{*}{616.6} & $O_1$ & 1516.3 & 10.6\ding{55} & 55.3\checkmark & 222.1 & 2.0 & 67.9  \\
&                        & $O_2$ & F      & 10.3\checkmark & 0    & 194.9 & 3.2 & 13.5  \\
&                        & $O_3$ & F      & 11.2\checkmark\checkmark & 0    & 204.2 & 3.1 & 14.3  \\
&                        & $O_4$ & F      & 11.3\checkmark & 0    & 270.7 & 3.2 & 14.5  \\
&                        & $O_5$ & F      & 11.6\checkmark & 0    & 234.0 & 3.2 & 14.8  \\
\midrule
\multirow{5}{*}{B2}
& \multirow{5}{*}{685.3} & $O_1$ & 1309.5 & 8.2\ding{55} & 40.7\checkmark & 196.9 & 1.2 & 50.1  \\
&                        & $O_2$ & F      & 8.1\ding{55} & 44.8\checkmark & 233.5 & 1.5 & 54.4  \\
&                        & $O_3$ & F      & 8.9\checkmark & 0    & 210.2 & 2.0 & 10.9  \\
&                        & $O_4$ & F      & 9.2\checkmark & 0    & 223.1 & 1.8 & 11.0  \\
&                        & $O_5$ & F      & 9.1\checkmark & 0    & 248.5 & 1.9 & 11.0  \\
\midrule
\multirow{5}{*}{MC}
& \multirow{5}{*}{796.3} & $O_1$ & 1114.0 & 11.5\ding{55} & Timeout(120)\ding{55} & 173.4 \checkmark& 16.7 & 321.6 \\
&                        & $O_2$ & F      & 12.2\checkmark & 0            & 203.2 & 16.7 & 28.9  \\
&                        & $O_3$ & F      & 11.9\checkmark & 0            & 211.3 & 16.9 & 28.8  \\
&                        & $O_4$ & F      & 11.3\checkmark & 0            & 239.3 & 17.0 & 28.3  \\
&                        & $O_5$ & F      & 12.1\checkmark & 0            & 226.4 & 17.0 & 29.1  \\
\midrule
\multirow{5}{*}{PD}
& \multirow{5}{*}{1259.5} & $O_1$ & F & 14.4\ding{55} & Timeout(240)\ding{55} & 637.1\checkmark & 40.8 & 932.3 \\
&                         & $O_2$ & F & 14.8\checkmark & 0            & 714.9 & 68.1 & 82.9  \\
&                         & $O_3$ & F & 14.5\checkmark & 0            & 667.0 & 71.8 & 86.3  \\
&                         & $O_4$ & F & 14.8\checkmark & 0            & 634.9 & 70.9 & 85.7  \\
&                         & $O_5$ & F & 14.1\checkmark & 0            & 699.6 & 70.0 & 84.1  \\
\bottomrule
\end{tabular}
}
\vspace{0.5ex}

\noindent\textbf{Remark.}
\textbf{Original BC}: time for synthesizing the original barrier certificate for $S_0$ and $S_u$;
\textbf{Global retrain}: time for retraining a policy and a global barrier certificate for $S_u$ and $S_u^{\mathrm{new}}$ from scratch;
\textbf{Stage III}: time for retraining a policy and a local barrier certificate for $S_u^{\mathrm{new}}$ in $S_{\mathrm{new}}$ from scratch;
\textbf{Verification}: time for computing the switching interface sets required for the compositional verification in \cref{theorem:apwbc};
\textbf{Total time}: end-to-end running time of our repair pipeline. ``F'' means global retraining failed within 7200s, and ``Timeout(x)'' means Stage~II failed within $x$ seconds.
\end{table*}

\begin{table*}[!htbp]
\centering
\caption{Running time (in seconds) under different scaling factors of repair regions.}
\label{tab:full-region-size}
\resizebox{\textwidth}{!}{
\begin{tabular}{c c c c c c}
\toprule
Benchmark 
& Scale of $S_{\mathrm{new}}$ 
& Stage I 
& Stage II 
& Stage III 
& Total time of Stages I,II \\
\midrule
\multirow{5}{*}{B1}
& $0.6\times$ & 5.6\ding{55}  & F\ding{55}    & F\ding{55}     & F    \\
& $0.8\times$ & 8.3\ding{55}  & F\ding{55}    & F\ding{55}     & F    \\
& $1.0\times$ & 11.0\checkmark & 0    & 204.2 & 11.0 \\
& $1.2\times$ & 17.2\checkmark & 0    & 281.5 & 17.2 \\
& $1.4\times$ & 19.9\ding{55} & 72.5\checkmark & 366.1 & 92.4 \\
\midrule
\multirow{5}{*}{B2}
& $0.6\times$ & 4.7\ding{55}  & F\ding{55}    & F\ding{55}     & F    \\
& $0.8\times$ & 7.0\ding{55}  & F \ding{55}   & F\ding{55}     & F    \\
& $1.0\times$ & 8.9\checkmark  & 0    & 210.2 & 8.9  \\
& $1.2\times$ & 12.5\checkmark & 0    & 270.9 & 12.5 \\
& $1.4\times$ & 15.1\ding{55} & 80.3\checkmark & 389.4 & 95.4 \\
\midrule
\multirow{5}{*}{MC}
& $0.6\times$ & 5.1\ding{55}  & F\ding{55}    & F\ding{55}     & F    \\
& $0.8\times$ & 8.2\checkmark  & 0    & 203.4 & 8.2  \\
& $1.0\times$ & 11.9\checkmark & 0    & 211.3 & 11.9 \\
& $1.2\times$ & 17.7\ding{55} & 77.3\checkmark & 363.3 & 95.0 \\
& $1.4\times$ & 22.3\ding{55} & F\ding{55}    & 402.5\checkmark & F    \\
\midrule
\multirow{5}{*}{PD}
& $0.6\times$ & 6.4\ding{55}  & F\ding{55}     & F\ding{55}     & F     \\
& $0.8\times$ & 10.5\checkmark & 0     & 642.9 & 10.5  \\
& $1.0\times$ & 14.5\checkmark & 0     & 667.0 & 14.5  \\
& $1.2\times$ & 22.0\ding{55} & 137.8\checkmark & 797.3 & 159.8 \\
& $1.4\times$ & 28.3\ding{55} & F\ding{55}     & 862.6\checkmark & F     \\
\bottomrule
\end{tabular}
}
\vspace{0.5em}

\parbox{0.96\textwidth}{
\textbf{Remark.}
\textbf{Stage III} denotes local retraining in the scaled repair region.
\textbf{Total time of Stages I,II} denotes the total running time of Stage~I and Stage~II for the current scaled region.
``F'' denotes failure due to timeout in the transfer stage.
}
\end{table*}

\begin{table}[!htbp]
\centering
\caption{Running time comparison of different template-selection strategies across four benchmarks.}
\label{tab:full-template-selection}
\resizebox{\textwidth}{!}{
\begin{tabular}{c c ccc ccc ccc}
\toprule
\multirow{2}{*}{Benchmark}
& \multirow{2}{*}{Obstacle ID}
& \multicolumn{3}{c}{Original global BC}
& \multicolumn{3}{c}{First repaired local BC}
& \multicolumn{3}{c}{Previous template} \\
\cmidrule(lr){3-5}
\cmidrule(lr){6-8}
\cmidrule(lr){9-11}
& 
& Stage I & Stage II & Step time
& Stage I & Stage II & Step time
& Stage I & Stage II & Step time \\
\midrule

\multirow{5}{*}{B1}
& O1
& 10.6 & 55.3 & 65.9
& -- & -- & --
& 10.6 & 55.3 & 65.9 \\
& O2
& 10.5 & 54.2 & 64.7
& 10.3 & 0 & 10.3
& 10.3 & 0 & 10.3 \\
& O3
& 10.9 & 0 & 10.9
& 10.6 & 0 & 10.6
& 11.0 & 0 & 11.0 \\
& O4
& 11.4 & 0 & 11.4
& 10.9 & 0 & 10.9
& 11.3 & 0 & 11.3 \\
& O5
& 11.3 & 0 & 11.3
& 11.3 & 0 & 11.3
& 11.6 & 0 & 11.6 \\

\midrule
\multirow{5}{*}{B2}
& O1
& 8.2 & 40.7 & 48.9
& -- & -- & --
& 8.2 & 40.7 & 48.9 \\
& O2
& 8.2 & 44.3 & 52.5
& 8.1 & 44.8 & 52.9
& 8.1 & 44.8 & 52.9 \\
& O3
& 8.7 & 0 & 8.7
& 8.6 & 0 & 8.6
& 8.9 & 0 & 8.9\\
& O4
& 9.0 & 0 & 9.0
& 9.3  & 0 & 9.3
& 9.2 & 0 & 9.2 \\
& O5
& 9.0 & 0 & 9.0
& 8.5 & 0 & 8.5
& 9.1 & 0 & 9.1 \\

\midrule
\multirow{5}{*}{MC}
& O1
&11.5& F & F
& -- & -- & --
&11.5& F & F   \\
& O2
& 11.6 & F & F
& 12.2 & 0 & 12.2
& 12.2 & 0 & 12.2 \\
& O3
& 12.1 & F & F
& 11.8 & 0 & 11.8
& 11.9 & 0 & 11.9 \\
& O4
& 11.7 & F & F
& 12.0 & 0 & 12.0
& 11.3 & 0 & 11.3 \\
& O5
& 11.7 & F & F
& 12.0 & 0 & 12.0
& 12.1 & 0 & 12.1 \\

\midrule
\multirow{5}{*}{PD}
& O1
& 14.4 & F & F
& -- & -- & --
& 14.4 & F & F \\
& O2
& 14.0 & F & F
& 14.8 & 0 & 14.8
& 14.8 & 0 & 14.8 \\
& O3
& 14.2 & F & F
& 14.0 & 0 & 14.0
& 14.5 & 0 & 14.5 \\
& O4
& 14.6 & F & F
& 14.3 & 0 & 14.3
& 14.8 & 0 & 14.8 \\
& O5
& 14.7 & F & F
& 14.2 & 0 & 14.2
& 14.1 & 0 & 14.1 \\

\bottomrule
\end{tabular}
}
\vspace{0.5em}
\begin{flushleft}
\footnotesize
\textbf{Remark.}
Step time is the total repair time at the current sequential step, i.e., the sum of Stage~I and Stage~II.
In the \emph{Previous repaired BC} strategy, the repaired local barrier certificate obtained at the previous step is used as the source template for the next step. "F" denotes failure due to timeout in the transfer stage. 
\end{flushleft}
\end{table}

\begin{table*}[!htbp]
\centering

\caption{Comparison between our method and shielding under multiple obstacles across benchmarks.}
\label{tab:full_shield_multi_obstacle}
\resizebox{0.8\textwidth}{!}{
\begin{tabular}{c c c c c c c c c}
\toprule
\multirow{2}{*}{Benchmark}
& \multirow{2}{*}{\# Obs.}
& \multicolumn{3}{c}{Ours}
& \multicolumn{3}{c}{Shield} \\
\cmidrule(lr){3-5} \cmidrule(lr){6-9}
&
& Total runtime
& Avg. reward
& Success rate (\%)
& \# Shield Interventions
& Total runtime
& Avg. reward
& Success rate (\%) \\
\midrule
\multirow{3}{*}{B1}
& $O_1$ & 67.9 & 57.10 & 100&622 & 269.8 & 56.14 & 100 \\
& $O_2$ & 81.4 & 55.37 & 100&972 & 413.6 & 53.96 & 100 \\
& $O_3$ & 95.7 & 54.59 & 100&1354 & 573.2 & 51.30 & 97  \\
\midrule
\multirow{3}{*}{B2}
& $O_1$ & 50.1  & 95.11 & 100&241 & 102.0 & 94.83 & 100 \\
& $O_2$ & 104.5 & 94.89 & 100& 270& 116.2 & 94.12 & 100 \\
& $O_3$ & 115.4 & 94.70 & 100&389 & 158.6 & 93.05 & 99  \\
\midrule
\multirow{3}{*}{MC}
& $O_1$ & 321.6 & 95.30 & 100&1265 & 587.6  & 93.19 & 100 \\
& $O_2$ & 350.5 & 93.63 & 100&1843 & 827.2  & 92.85 & 100 \\
& $O_3$ & 379.3 & 93.55 & 100&2360 & 1010.0 & 92.73 & 100 \\
\midrule
\multirow{3}{*}{PD}
& $O_1$ & 932.3  & $-155.24$ & 100&2428 & 1403.0 & $-168.1$ & 100 \\
& $O_2$ & 1015.2 & $-162.53$ & 100&4306 & 1666.2 & $-173.7$ & 100 \\
& $O_3$ & 1101.5 & $-175.92$ & 100&5789 & 1841.0 & $-199.5$ & 100 \\
\bottomrule
\end{tabular}
}
\vspace{0.5ex}
\footnotesize

\noindent\textbf{Remark.}
\textbf{Total runtime}: total runtime in seconds over the 200 runs;
\textbf{Avg. reward}: the mean cumulative reward;
\textbf{Success rate}: percentage of successful runs;
\textbf{\# Shield Interventions}: number of times that the shielder  modifies or replaces the nominal action across the 200 runs.

\end{table*}

\begin{table*}[!htbp]
\centering
\caption{Effect of the learning rate on global retraining.
Numerical entries report the mean end-to-end runtime
(in seconds) over 20 runs per configuration.}
\label{tab:retrain_learning_rate}

\begin{tabular}{c c c c c}
\toprule
\multirow{2}{*}{Benchmark}
& \multirow{2}{*}{Obstacle ID}
& \multicolumn{3}{c}{Learning rate} \\
\cmidrule(lr){3-5}
&
& $10^{-5}$ (main setting)
& $10^{-4}$
& $10^{-3}$ \\
\midrule

\multirow{5}{*}{B1}
& $O_1$ & 1728.5 & F & F \\
& $O_2$ & F & F & F \\
& $O_3$ & F & F & F \\
& $O_4$ & F & F & F \\
& $O_5$ & F & F & F \\
\midrule

\multirow{5}{*}{B2}
& $O_1$ & 1469.3 & F & F \\
& $O_2$ & F & F & F \\
& $O_3$ & F & F & F \\
& $O_4$ & F & F & F \\
& $O_5$ & F & F & F \\
\midrule

\multirow{5}{*}{MC}
& $O_1$ & 1331.2 & F & F \\
& $O_2$ & F & F & F \\
& $O_3$ & F & F & F \\
& $O_4$ & F & F & F \\
& $O_5$ & F & F & F \\
\midrule

\multirow{5}{*}{PD}
& $O_1$ & F & F & F \\
& $O_2$ & F & F & F \\
& $O_3$ & F & F & F \\
& $O_4$ & F & F & F \\
& $O_5$ & F & F & F \\
\bottomrule
\end{tabular}

\vspace{0.5ex}
\begin{minipage}{\textwidth}
\footnotesize
\textbf{Remark.}
 ``F'' means global retraining failed within 7200s.

\end{minipage}
\end{table*}

\FloatBarrier
\newpage
\end{document}